\documentclass{article} % For LaTeX2e
\PassOptionsToPackage{square,numbers}{natbib}

\usepackage[preprint]{neurips_2023}

\usepackage[utf8]{inputenc} % allow utf-8 input
\usepackage[T1]{fontenc}    % use 8-bit T1 fonts
\usepackage{hyperref}       % hyperlinks
\usepackage{url}            % simple URL typesetting
\usepackage{booktabs}       % professional-quality tables
\usepackage{amsfonts}       % blackboard math symbols
\usepackage{nicefrac}       % compact symbols for 1/2, etc.
\usepackage{microtype}      % microtypography
\usepackage{xcolor}         % colors
\usepackage{graphicx}
\usepackage{float}
\usepackage{amsmath,amssymb,amsfonts}
\usepackage{mwe}
\usepackage{subfig}
\usepackage{svg}

\usepackage{amsmath}
\usepackage[makeroom]{cancel}
\usepackage[english]{babel}
\title{Estimating Post-Synaptic Effects for Online Training of Feed-Forward SNNs}

\author{%
  Thomas M. Summe \\
  Dept. of Computer Science and Engineering\\
  University of Notre Dame\\
  \texttt{tsumme@nd.edu} \\
  \And
  Clemens JS Schaefer \\
  Dept. of Computer Science and Engineering\\
  University of Notre Dame \\
  \texttt{cschaef6@nd.edu} \\
  \And
  Siddharth Joshi \\
  Dept. of Computer Science and Engineering\\
  University of Notre Dame \\
  \texttt{sjoshi2@nd.edu} \\
  }

\begin{document}

\maketitle

\begin{abstract}
Facilitating online learning in spiking neural networks (SNNs) is a key step in developing event-based models that can adapt to changing environments and learn from continuous data streams in real-time. Although forward-mode differentiation enables online learning, its computational requirements restrict scalability. This is typically addressed through approximations that limit learning in deep models. In this study, we propose Online Training with Postsynaptic Estimates (OTPE) for training feed-forward SNNs, which approximates Real-Time Recurrent Learning (RTRL) by incorporating temporal dynamics not captured by current approximations, such as Online Training Through Time (OTTT) and Online Spatio-Temporal Learning (OSTL). We show improved scaling for multi-layer networks using a novel approximation of temporal effects on the subsequent layer's activity. This approximation incurs minimal overhead in the time and space complexity compared to similar algorithms, and the calculation of temporal effects remains local to each layer. We characterize the learning performance of our proposed algorithms on multiple SNN model configurations for rate-based and time-based encoding. OTPE exhibits the highest directional alignment to exact gradients, calculated with backpropagation through time (BPTT), in deep networks and, on time-based encoding, outperforms other approximate methods. We also observe sizeable gains in average performance over similar algorithms in offline training of Spiking Heidelberg Digits with equivalent hyper-parameters (OTTT/OSTL – 70.5$\%$; OTPE – 75.2$\%$; BPTT – 78.1$\%$).

\end{abstract}

\section{Introduction}

% Spiking neural networks (SNNs) have been introduced as the third generation of artificial neural networks, allowing theoretical energy savings and greater alignment with biology \cite{}. Unlike traditional deep neural networks, SNNs produce sparse, binary outputs, which reduces the computational cost of inference. Training these models, however, remains a challenge. Among multiple challenges is online training, which enables neural networks to learn and operate on streaming data.

% The defacto training method of SNNs is via backpropagation through time (BPTT). By storing the output of each layer at every time-step, the network is unrolled through time for gradient calculation. This process, however, is impractical for online learning and scales memory costs by the number of time-steps. An alternative exact gradient calculation method that addresses these issues is real-time recurrent learning (RTRL).  RTRL computes Jacobian-vector products via forward-mode differentiation, not requiring scaling storage by time or unrolling. These intermediate gradients, however, are too large for training all but the smallest networks. 

%  must be used to enable online learning for SNNs

Spiking neural networks (SNNs) promise a path toward energy-efficient machine intelligence for streaming data~\cite{yik2023neurobench}. Despite this potential, efficiently training them on temporal sequences remains a challenge. Backpropagation through time (BPTT) applied with surrogate gradient estimates of SNN neurons~\citep{neftci2019surrogate,superspike,slayer} has become the dominant method to train SNNs. However, BPTT is unsuitable for online learning  \citep{neftcidecolle,bohnstingl2022online}

Real-time recurrent learning (RTRL)~\cite{williams1989experimental} computes exact gradients for stateful models without temporal unrolling. This enables online learning at the cost of $O(n^3)$ storage and $O(n^4)$ compute, which is not practical for training all but the smallest networks. To address these limitations, practical implementations approximate RTRL by limiting parameter influence temporally or spatially, adopting low-rank matrix approximations, leveraging stochastic gradient estimates, incorporating model-specific assumptions, or selectively omitting certain influence pathways \citep{mujika2018approximating,benzing2019optimal,tallec2018unbiased,bohnstingl2022online,menick2020practical}. Of these, a promising approach for SNN training involves storing only the influence matrix to gradient values along the diagonal of the stored Jacobians. This approximation restricts temporal influence of the gradient calculations to the current output from a single layer, which neglects how the previous outputs of a neuron impact the membrane potential of downstream neurons. This approximation reduces learning performance when compared to the exact gradient calculations in RTRL or BPTT \cite{bohnstingl2022online}, due to untracked causal effects over time. To address this, we develop a novel approximation of this temporal effect through our algorithm, Online Training with Postsynaptic Estimates (OTPE). OTPE maintains a trace of parameter influence over multiple time-steps, implementing a comprehensive approximation of the entire temporal effect in a one-hidden-layer model. In deep networks, spikes generated in earlier layers will have a delayed influence on spiking activity of neurons in deeper layers. Thus, gradient approximation worsens as the error is back-propagated if only the current time-step is considered. Intuitively, the performance difference between Online Spatio-Temporal Learning (OSTL) \citet{bohnstingl2022online} and BPTT reflects the impact of these residual temporal effects since the explicit difference between OSTL and exact gradient calculation is the exclusion of these residual effects.
%\sj{we should talk about why this matters for a deep network, in other words, we get exact results for a single-hidden-layer model but what can we say about deeper models}.
% \ts{we should defer it to the algorithm description if anything} \sj{we need to be more specific here, what conditions? Does it make sense to talk about this right now or defer this to the detailed sections?}
We demonstrate that OTPE's approximation significantly increases alignment with exact gradients. We observe a $\sim70\%$ increase in gradient cosine similarity to BPTT, in the first hidden layer and $\sim50\%$ in the second hidden layer of a 2-hidden-layer network for a model trained on the Spiking Heideberg Digits (SHD) dataset \citep{cramer2020heidelberg}. We consistently observe similar improvements for both online and offline learning, and across other evaluated datasets and model configurations. %\sj{what is "significantly increase" here, quantify this.}.

Our primary contributions include: \vspace{-.2cm}

 \begin{itemize}\setlength\itemsep{0em}% [topsep=0pt,itemsep=-1ex,partopsep=1ex,parsep=1ex]

    \item a novel approximation of RTRL, OTPE, that captures the effects of multi-step temporal sequences through a spiking neuron which are excluded from previous algorithms; % effects neglected by other algorithms without reducing scalability or locality and demonstrate its superior temporal learning compared to the other approximate methods.
    \item a further relaxation of our algorithm that can achieve similar scalability to state-of-the-art while delivering superior learning performance on multiple tasks;
    \item in-depth evaluations of OTPE against existing SNN training algorithms, including gradient approximation quality and learning performance in temporal and rate-encoded inputs. %We evaluate the quality of existing online learning gradient approximations on temporal learning directly (through cosine similarity) and indirectly (through task accuracy). We also qualitatively evaluate them through their loss landscape.    
\end{itemize}

\section{Background}

\begin{figure}
% \begin{minipage}{.5\columnwidth}
\centering
\subfloat[]{\label{main:a}\includegraphics[width=.4\columnwidth]{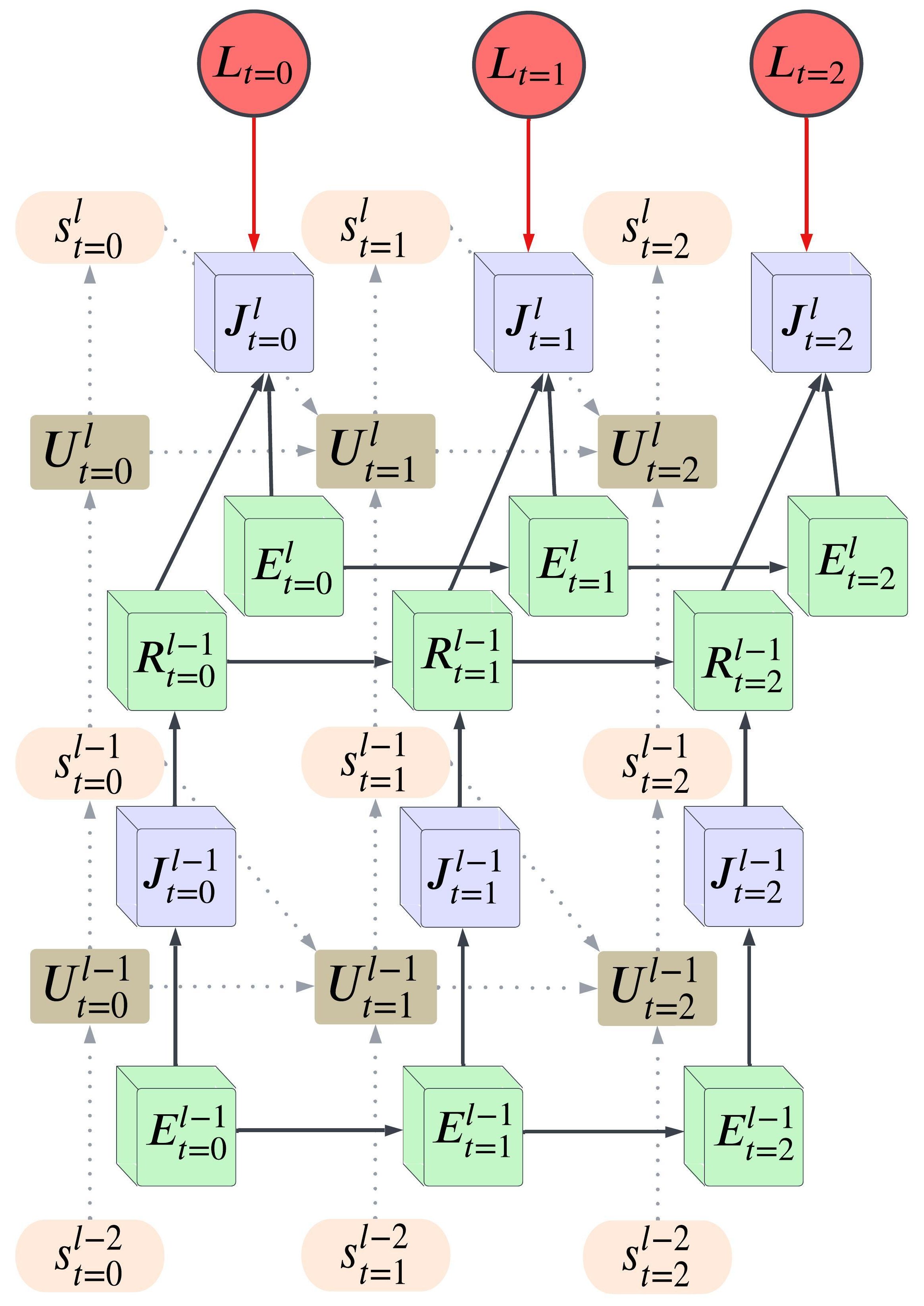}}\hfill
% \centering
\subfloat[]{\label{main:b}\includegraphics[width=.4\columnwidth]{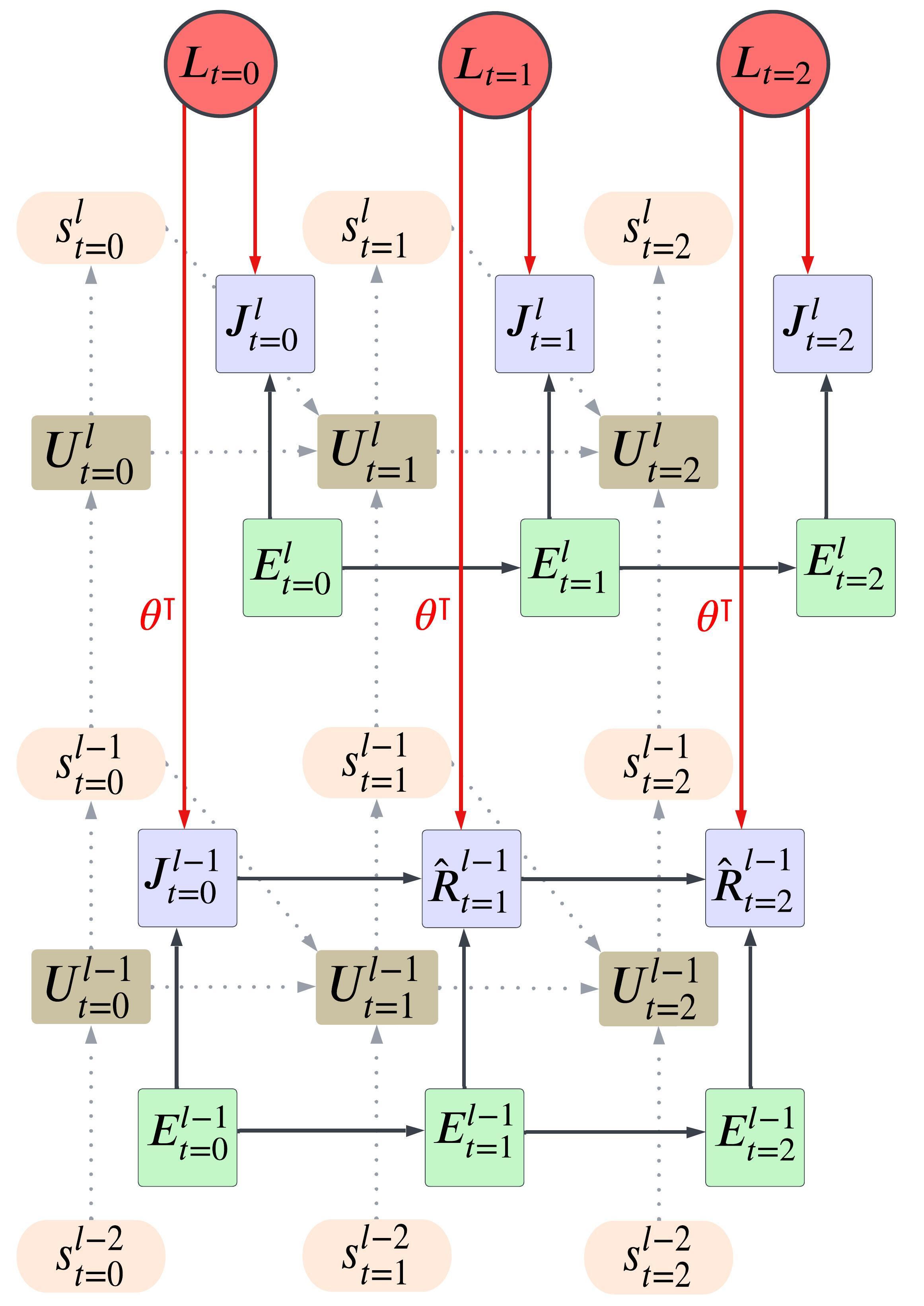}}\vspace{-.3cm}
\subfloat{\label{main:legend}\includegraphics[trim={0 0 0 2cm},clip, width=.8\columnwidth]{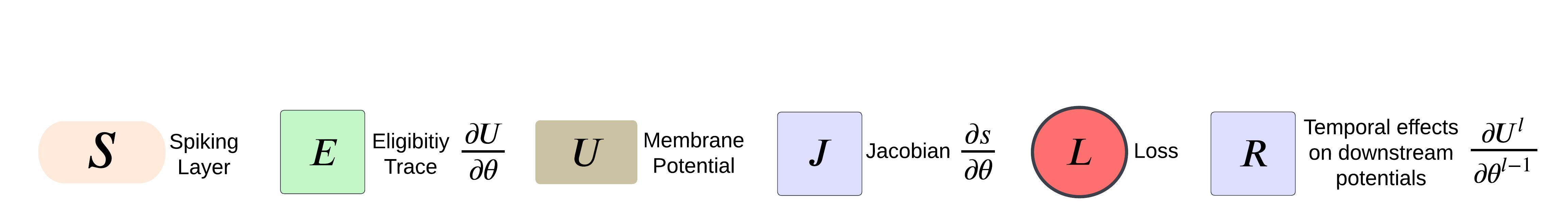}}
\caption{Depiction of RTRL and OTPE. The grey dotted lines indicate the forward pass effects through a feed-forward spiking neural network. (a) A graphical depiction of computations incurred in RTRL. Here, the Jacobian ($J$) matrix is calculated for all the upstream parameters in the model. Thus, we represent this by gradients in cubes, reflecting the $n^3$ size of the Jacobian. (b) Depiction of OTPE. Unlike RTRL, all temporal gradient calculations are layer-local, and loss is backpropagated.}
\label{fig:foward-pass}\vspace{-.3cm}
\end{figure}

Training stateful networks such as SNNs requires temporal credit assignment~\cite{pertlearn,randomfeedback,bohnstingl2022online,xiao2022online,neftcidecolle}. Of these different methods, state-of-the-art results in SNN training typically employ BPTT-based credit assignment \cite{eshraghian2023training} with surrogate derivatives to calculate gradients through the SNNs discontinuity \citep{zenke2021remarkable,superspike}. Consequently, we compare the gradient approximation quality of OTPE, Online Training Through Time (OTTT)\citet{xiao2022online}, and OSTL against the exact gradients computed by BPTT. All methods are applied to deep feed-forward SNNs composed of leaky integrate-and-fire (LIF) neurons in fully connected layers.

% However, because gradient-based algorithm (BPTT) is unsuited to online learning. Consequently, we study the gradient-calculations and the resultant model performance for our algorithms and OTTT and OSTL, compared against BPTT.

% in addition to comparing results against BPTT, we also comapre our work to other recent online learning algorithms targeting SNN training?

% BPTT with surrogate gradients as the reference. We additionally compare OTPE against other recent reverse-auto

% In this study, we implement and evaluate four algorithms and compare them against BPTT. These are OSTL, OTTT, OTPE, and Approximate OTPE.
%  discrete, and therefore, do not provide useful derivatives, surrogate derivatives of the Heaviside step function provide useful approximate gradients for training \ts{citation needed}
% Among these, training algorithms like Perturbation Learning, Random Feedback, Local Losses, etc., have seen wide exploration for SNNs~\cite{pertlearn,randomfeedback,neftcidecolle}.
%  which utilizes random weight perturbations to gauge error changes, and  positing that different neural layers could be guided by distinct, localized cost functions. Such methods aim to address the inherent challenges SNNs present, be it their complex neuronal dynamics or the spatial credit assignment problem.

\subsection{LIF neuron}

SNNs promise low computational requirements, arising from activation sparsity and unary outputs. LIF neurons are the most commonly used for balancing performance and complexity, especially in deep models \cite{superspike}. Similar to a plethora of other work~\cite{fang2021incorporating}, we use a subtraction-based reset formulation of the LIF, with the neuron behavior written as
\begin{align}
    s^l_t &= H(\lambda U^l_{t-1} + s^{l-1}_t \cdot \theta - V_{th}), \nonumber \\ 
    U^l_t &= \lambda U^l_{t-1} - V_{th} \cdot s^l_t.\nonumber
\end{align}
Here, $U^l_t$ is the neuron's membrane potential in layer $l$ at time-step $t$, which decays by the leak $\lambda$ while accumulating spiking inputs $s^{l-1}_t$ from the previous layer, weighted by $\theta$. The neuron emits a spike $s^{l}_t$ whenever its membrane potential exceeds the threshold $V_{th}$. The derivative of the Heaviside step function ($H$) is the Dirac delta function which is zero almost everywhere, effectively setting all gradients to zero. To generate non-zero gradients, we employ surrogate gradients, which replace the dirac delta function with the derivative of the fast sigmoid function~\cite{neftci2019surrogate}.

\subsection{RTRL}

Real-Time Recurrent Learning (RTRL) (Fig.~\ref{fig:foward-pass}) calculates gradients through time using forward-mode differentiation, calculating and storing Jacobian-vector products. While BPTT must store outputs at each layer and unroll the network to perform reverse-mode differentiation through time, RTRL stores and updates each parameter's effects on the network's state. Because the stored Jacobian tracks every parameter's influence on each state variable ($O(n^3)$ in the number of parameters) the network avoids unrolling to calculate gradients. Due to this, RTRL can calculate exact gradients for online learning. RTRL gradient calculation for the output layer of an SNN can be written as
\begin{equation} \label{eq:RTRL-1}
\begin{split}
\frac{\partial \mathcal{L}}{\partial \theta^{l}}& =
\sum_{t=1}^T \frac{\partial \mathcal{L}^l_t}{\partial s^l_t} 
\frac{\partial s^l_t}{\partial U^l_t}
\frac{\partial U^l_t}{\partial \theta^{l}} \\ & =
\sum_{t=1}^T 
\frac{\partial \mathcal{L}_t}{\partial s^l_t}
\frac{\partial s^l_t}{\partial U^l_t}
\left(\frac{\partial U^l_t}{\partial \theta^{l}_t}+
\frac{\partial U^l_t}{\partial U^l_{t-1}} 
\frac{\partial U^l_{t-1}}{\partial \theta^{l}}\right).
\end{split}
\end{equation}
We denote the loss with $\mathcal{L}$, the spike output with $s$,  the membrane potential with $U$, the parameters with $\theta$, the total number of time-steps with $T$, and the current time-step with $t$.
We can recursively calculate and store the temporal gradients, $\frac{\partial U^l_t}{\partial \theta^l}$, in eqn~\eqref{eq:RTRL-1} through
% \begin{equation}
$\frac{\partial U^l_t}{\partial \theta^l} = 
\left(\frac{\partial U^l_t}{\partial \theta^l_t} +
\frac{\partial U^l_t}{\partial U^l_{t-1}}
\frac{\partial U^l_{t-1}}{\partial \theta^l}\right)$.
%\nonumber
%\end{equation}

For the hidden layer, we can expand eqn~\eqref{eq:RTRL-1}, substituting $\frac{\partial U^l_t}{\partial \theta^l}$ with $\frac{\partial U^l_{t}}{\partial \theta^{l-1}_t}$, resulting in
\begin{equation} \label{eq:RTRL-2_expand}
\begin{split}
\frac{\partial U^l_{t}}{\partial \theta^{l-1}_t}& =
\frac{\partial U^l_t}{\partial s^{l-1}_t}
\frac{\partial s^{l-1}_t}{\partial \theta^{l-1}_t} \\ & = 
\frac{\partial U^l_t}{\partial s^{l-1}_t}
\left(
\frac{\partial s^{l-1}_t}{\partial U^{l-1}_t}
\left(
\frac{\partial U^{l-1}_{t}}{\partial \theta^{l-1}_t} +
\frac{\partial U^{l-1}_t}{\partial U^{l-1}_{t-1}}
\frac{\partial U^{l-1}_{t-1}}{\partial \theta^{l-1}}
\right)
\right).
\end{split}
\end{equation}

Where, the kernel's transpose in a dense layer, $\theta^\intercal$, is given by $\frac{\partial U^l_t}{\partial s^{l-1}_t}$.

\subsection{Practical Approaches to RTRL}
Prior work achieves online capabilities by approximating exact gradient computation. Of particular interest are OSTL \cite{bohnstingl2022online} and OTTT \cite{xiao2022online}. OSTL focuses on achieving bio-plausibility by implementing eligibility traces \cite{gerstner2018eligibility}, through a mechanism derived from RTRL. Another similar algorithm is OTTT, which ignores the reset mechanism in the gradient calculation of SNNs to reduce the storage and compute overhead relative to OSTL.

\subsubsection{OSTL}

OSTL is an RTRL approximation that separates spatial and temporal gradients for calculating the overall gradients. Because $\frac{\partial U_t}{\partial \theta}$ in RTRL has $n^3$ entries (where n is the layer size), the storage demand of RTRL limits scalability. OSTL approximates $\frac{\partial U_t}{\partial \theta}$ by assuming that all nonzero elements are along the diagonal, thus reducing its size to $n^2$. This assumption holds for a feed-forward SNN, achieving exact gradient computation in a network without hidden layers.

In order to train a deep network, OSTL backpropagates a spatial gradient through the current time-step ($t$) and combines this with a temporal gradient, which maintains how a layer's parameters influenced its most recent output, $s_t$. While all temporal dynamics concerning a single layer's influence on $s_t$ are accounted for, the influence of any spiking activity at previous time-steps is excluded, as shown
\begin{equation}
\frac{\partial \mathcal{L}}{\partial \theta^{l-1}} =
\sum_{t=1}^T \frac{\partial \mathcal{L}^l_t}{\partial s^l_t} 
\frac{\partial s^l_t}{\partial U^l_t}
\frac{\partial U^l_t}{\partial \theta^{l-1}} =
\sum_{t=1}^T 
\frac{\partial \mathcal{L}_t}{\partial s^l_t}
\frac{\partial s^l_t}{\partial U^l_t}
\left(\frac{\partial U^l_t}{\partial \theta^{l-1}_t}+
\xcancel{\frac{\partial U^l_t}{\partial U^l_{t-1}} 
\frac{\partial U^l_{t-1}}{\partial \theta^{l-1}}}\right).\nonumber
\end{equation}

\subsubsection{OTTT}

OTTT is conceptually similar to OSTL. Since $\frac{\partial s^l_t}{\partial U^l_t}$ is the Dirac delta function, a surrogate derivative normally takes its place, but OTTT chooses only to apply the surrogate derivative when calculating the spatial gradient and refrains from doing so when updating the temporal gradient. In other words, the gradient calcuation from BPTT which uses $s^{l-1}_t$ can be substituted in for the calculation in RTRL for $\frac{\partial U^l_t}{\partial \theta^{l}_t}$.  Since the derivative of the Heaviside step function is zero almost everywhere, the temporal gradient in a subtraction-based LIF neuron simplifies to the summation over time of the product between the time-weighted leak $\lambda^{T-t}$ and $s^{l-1}_t$. Consequently, only a running weighted sum of the input, $\hat{a}$ in \cite{xiao2022online}, is required to be stored and updated, reducing the space complexity of OTTT to $O(n)$.
%  $\frac{\partial U^l_t}{\partial \theta^{l}}$, rewritten then as . Since $\frac{\partial U^l_t}{\partial \theta^{l}_t}$ is equivalent to $s^{l-1}_t$,
While the surrogate derivative is normally applied as $\frac{\partial U_t}{\partial U_{t-1}} = \lambda + \frac{\partial U_t}{\partial s_t}
\frac{\partial s_t}{\partial U_{t-1}}$. OTTT uses the Heaviside function instead, such that $\frac{\partial U_t}{\partial U_{t-1}} =
\lambda +
\frac{\partial U_t}{\partial s_t} \cdot
\emptyset$, which is simply $\lambda$. 
Consequently, OTTT's temporal gradient, $\hat{a}$, can be calculated as
\begin{equation}
\hat{a}^{l}_{t=T} =
\sum_{t=1}^T \lambda^{T-t} s^{l-1}_t.
\end{equation}

% \begin{equation}
% \frac{\partial U_t}{\partial U_{t-1}} =
% \lambda +
% \frac{\partial U_t}{\partial s_t}
% \frac{\partial s_t}{\partial U_{t-1}}\nonumber.
% \end{equation}

% \begin{equation}
% \frac{\partial U_t}{\partial U_{t-1}} =
% \lambda +
% \frac{\partial U_t}{\partial s_t} \cdot
% \emptyset =
% \lambda.\nonumber
% \end{equation}

\section{Postsynaptic Estimation for SNNs}
% \subsection{OTPE}
In a feed-forward SNN with hidden layers, temporal dynamics that influence subsequent layers in the network are not addressed by current approximations. We achieve a space-efficient method of approximating these gradients in a similar fashion to OTTT.

Specifically, we do not apply the surrogate derivative for $\frac{\partial s^{l+1}_t}{\partial U^{l+1}_t}$ when calculating $\frac{\partial U^{l+1}_{t+1}}{\partial U^{l+1}_t}$ during the calculation of the temporal dynamics in layer $l$. We also assume the time constant is global. During forward gradient computation in layer $l$, we assume the subsequent layer's temporal dynamics are captured by the running sum of the subsequent layer's inputs, which are the output spikes of layer $l$. By maintaining a running weighted sum of how the parameters in layer $l$ influenced previous spikes, the parameters' contribution to the subsequent layer's temporal dynamics is represented during gradient calculation. In order to calculate how the parameters affect spikes at the current time-step, OTPE implements OSTL to produce the diagonal Jacobian matrix, $\frac{\partial{s(t)^l_i}}{\partial w^l_{ij}}$ at each time-step.

OTPE selectively applies the surrogate derivatives such that
\begin{equation} \label{eq:OTPE}
\frac{\partial \mathcal{L}}{\partial \theta^{l-1}} =
\sum_{t=1}^T \frac{\partial \mathcal{L}^l_t}{\partial s^l_t} 
\frac{\partial s^l_t}{\partial U^l_t}
\frac{\partial U^l_t}{\partial \theta^{l-1}} =
\sum_{t=1}^T 
\frac{\partial \mathcal{L}_t}{\partial s^l_t}
\frac{\partial s^l_t}{\partial U^l_t}
\left(\frac{\partial U^l_t}{\partial \theta^{l-1}_t}+
\lambda \cdot
\frac{\partial U^l_{t-1}}{\partial \theta^{l-1}}\right).
\end{equation}

Since $\frac{\partial U^l_t}{\partial s^{l-1}_t}$ in eqn~\eqref{eq:RTRL-2_expand} is the kernel's transpose, $\theta^\intercal$, we avoid recursion, and can write
\begin{equation} \label{eq:OTPE_expand}
\frac{\partial U^l_{t=T}}{\partial \theta^{l-1}} =
{\theta^{l}}^\intercal
\sum_{t=1}^T \lambda^{T-t} 
\frac{\partial s^{l-1}_t}{\partial \theta^{l-1}} =
{\theta^{l}}^\intercal
\hat{R}.
\end{equation}
As seen in eqn~\eqref{eq:OTPE_expand}, similar to OSTL, we only need to maintain a running weighted sum of $\frac{\partial s^{l-1}_t}{\partial \theta^{l-1}}$. % which we calculate in the same manner as OSTL, in order to find how the parameters of a hidden layer affect the membrane potential of the following layer.

\subsection{Approximate OTPE}

Although OTPE achieves scalability comparable to OSTL, OTTT remains more scalable. To combine the benefits of OTPE's temporal information with $O(n)$ space complexity, we approximate our temporal calculation in OTPE to only store vectors of size n, by assuming the same temporal dynamics as OTTT for a single layer. We then store and update a weighted sum of OTTT's weighted sum ($\hat{a}$) which we call $\hat{z}$. We additionally maintain a running weighted average of the surrogate gradients through time, which we refer to as $\bar{g}$. When the spatial gradient reaches a layer, $\bar{g}$ is used in place of the immediate time-step's surrogate. The kernel gradients are calculated by taking the outer product of the back-propagated loss and $\hat{z}$.

\begin{equation} \label{eq:Approx_OTPE}
\frac{\partial \mathcal{L}^l_T}{\partial \theta^{l-1}} \approx
\left( \left( \frac{\partial \mathcal{L}^l_T}{\partial s^{l}} \cdot {\theta^{l}}^\intercal \right) \cdot \bar{g} \right)
\otimes \hat{z}
\end{equation}

Intuitively, this approximation relies on the viewing $\hat{R}$ in OTPE as a second-order filter of the inputs. If the reset mechanism is ignored, the eligibility trace in OTPE is simply a running weighted sum of the inputs. Unlike OTTT, however, the second-order filter does not correspond to the surrogate derivative at only the current time-step. To maintain $O(n)$ space complexity, we assume the membrane potential's influence on the spike output at each time-step is approximately equivalent to the running weighted average $\bar{g}$. Gradient backpropagation for OTPE or Approx OTPE is similar to OSTL and OTTT. With the exception of using $\bar{g}$ instead of the most recent time-step's surrogate gradients.

\subsection{F-OTPE}

The formulation of OTPE in eqn~\eqref{eq:OTPE} exactly calculates the gradients of the hidden layer in a 1-hidden-layer feed-forward SNN. However, $\hat{R}$ can also be applied to an accumulating weighted sum of the output layer's spikes. This provides an additional capability to calculate loss based on the history of model outputs rather than applying the loss function only to the most recent output. Unlike the reset mechanism in a neuron, the accumulation of the network's output is linear and, therefore, accurately captured by $\hat{R}$, with a given leakage term on the accumulating output. By doing this, the loss can be calculated similarly to how cross-entropy loss is calculated in offline training. We test the application of OTPE to all layers in the network for online learning, applying cross-entropy loss to a leaking sum of the model's output. We also do this for its approximation, denoting both with ``F-''. 

Table~\ref{tab:complexity}, summarizes the time and space complexity of the evaluated algorithms.
\begin{table}[]
  \centering
  \begin{tabular}{lcc}
    \cmidrule(r){1-3}
    Name     & Space Complexity     & Time Complexity \\
    \midrule
    BPTT & $O(Tn)$ & $O(Tn^2)$ \\
    OTTT & $O(n)$ & $O(Tn + n^2)$\\
    OSTL& $O(n^2)$ & $O(Tn^2)$ \\
    \bottomrule
 Approx OTPE & $O(n)$ &$O(Tn + n^2)$\\
 OTPE & $O(n^2)$ &$O(Tn^2)$ \\
  \end{tabular}
  \caption{Time and space complexity of BPTT and all tested approximate algorithms for calculating gradients at time-step $T$. The complexity is in reference to a single dense layer with an input and output size $n$ and batch size of 1.}\label{tab:complexity}
\end{table}

\section{Experiments}
% \label{headings}
We test and compare all algorithms for online and offline training on Randman \citep{zenke2021remarkable}, a synthetic spiking dataset, and SHD \citep{cramer2020heidelberg} (dataset related parameters provided in \ref{ap:hyperparam1}). In offline learning, we evaluate accuracy, cosine similarity to exact gradients produced via BPTT, and training trajectories in the loss landscape. In offline training on SHD, we evaluate test accuracy across multiple model configurations to identify suitable hyperparameters for comparing all algorithms. For online training on SHD, we run a learning rate search across three model configurations for the same purpose (see Appendix \ref{ap:HP_search} for full hyperparameter search results). All tests were performed on NVIDIA GPUs, using the Adamax optimizer  \cite{kingma2014adam}. We chose the Adamax optimizer because of its relative effectiveness on temporal learning and use in the original SHD study. We provide JAX code to enable reproduction of our results~\cite{jax2018github,flax2020github}. For all tests, we use a fast sigmoid slope of 25, the default value in snnTorch \cite{eshraghian2023training}. BPTT also delivered highest accuracy with this value in our hyperparameter search (Appendix \ref{ap:HP_search})

\paragraph{Randman} is a synthetic dataset, aimed at studying SNN capabilities in learning spike timing patterns. Spike-times are generated on a random smooth manifold, projected onto a high-dimensional hypercube. These are then sampled to generate a spike-train, which the SNN learns to classify. The dimensionality and smoothness of the manifold is varied to adjust the task difficulty. Neurons, firing once per trial, only contain temporal information, a format we term T-Randman. We also modify Randman to test rate-based learning through R-Randman. Here, spike-rates are determined by manifold values and are generated to be temporally uncorrelated through random shuffling.

% spike trains  consists of a set of input spike trains that are generated by projecting a random smooth manifold onto a high-dimensional hypercube. Data is produced by sampling spike times on the manifold, and the SNN learns to classify it. The dimensionality and smoothness of the manifold can be varied to adjust the task difficulty. Randman allows us to separate spike-timing temporal learning from rate-based learning since, by default, each neuron only spikes once per trial, thus containing no rate-based information. We denote this time-encoded form of Randman as T-Randman. To test rate-based learning, we modify Randman so the value sampled from the manifold determines the number of spikes that occur rather than when a spike occurs. The spikes are shuffled across the time-steps so there is no temporal correlation to their occurrence. We denote this rate-encoded form of Randman as R-Randman. \ts{See appendix for details on Randman data generation.}

\paragraph{Spiking Heidelberg Digits (SHD)} is a spiking dataset consisting of 10,000 recordings of spoken digits 0-9 in German and English, across 20 classes. The recordings are processed through a model of the inner ear to produce a spiking representation. Since Randman's structure does not reflect natural data, we also evaluate model performance on SHD to reflect performance for a practical application. We evaluate accuracy after both online and offline training for SHD.% allows us to measure temporal learning exclusively, these results do not directly speak to performance gains in real datasets. We evaluate the test accuracy after online and offline training for SHD.

% All algorithms are implemented using JAX \cite{jax2018github}, Flax \cite{flax2020github}, and Optax. All tests were performed on NVIDIA GPUs, using the Adamax \cite{kingma2014adam} optimizer. We chose the Adamax optimizer because of its relative effectiveness on temporal learning and use in the original SHD study.

\subsection{Evaluating Learning Performance}

\begin{figure}[tbp]
\centering
\subfloat[]{
\includegraphics[width=.31\textwidth]{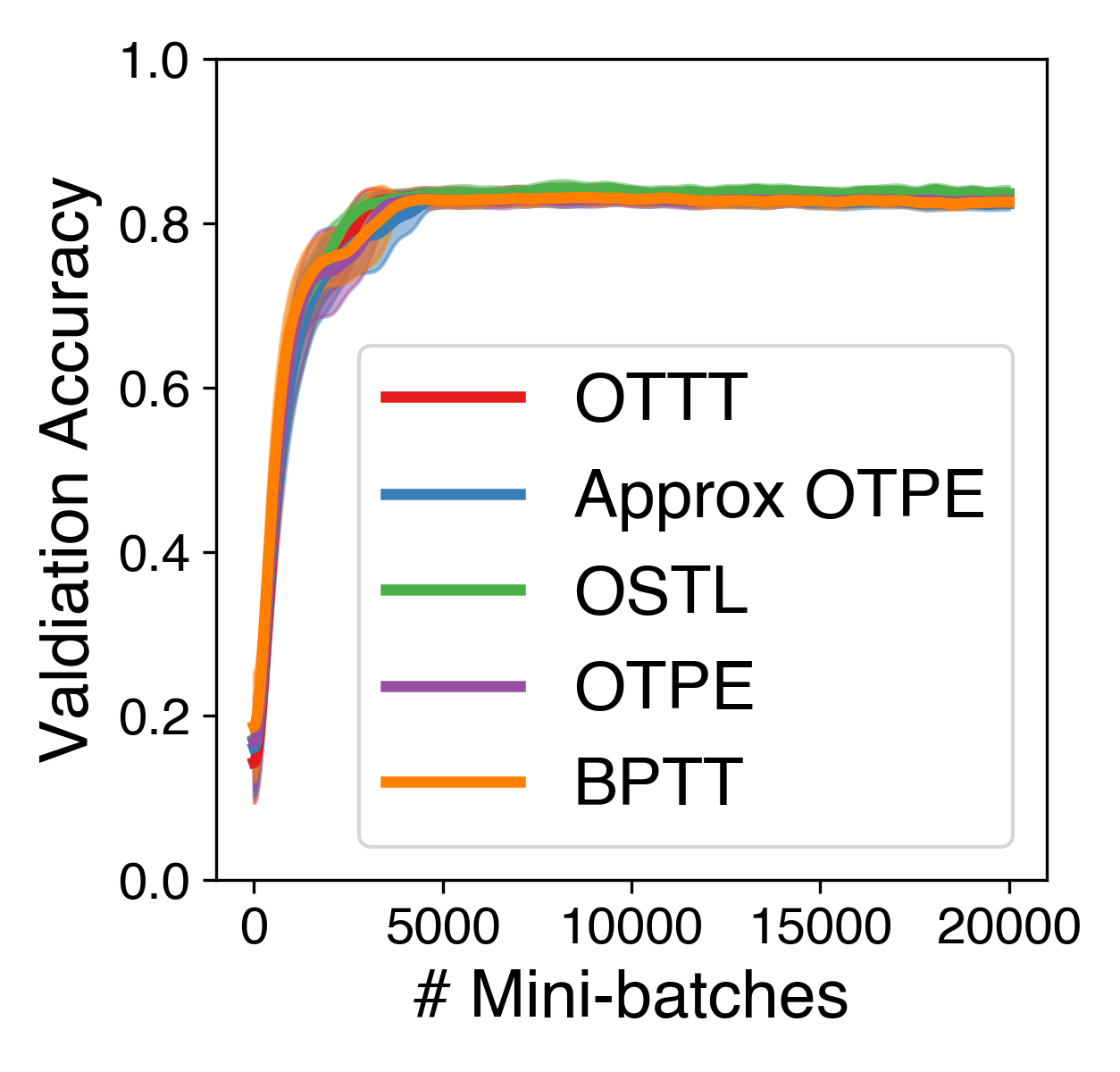}\hfill}
\subfloat[]{
\includegraphics[width=.30\textwidth]{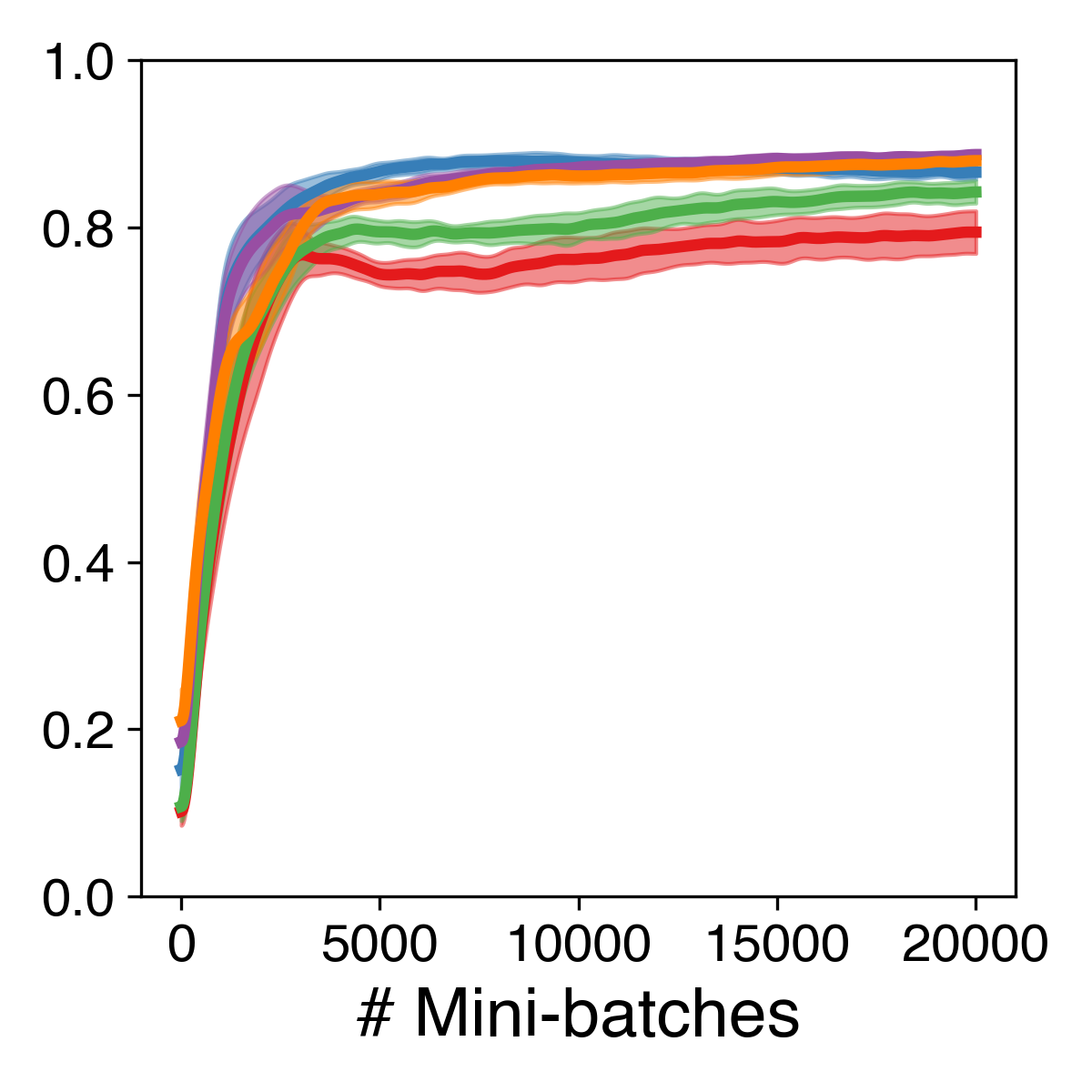}\hfill}
\subfloat[]{
\includegraphics[width=.30\textwidth]{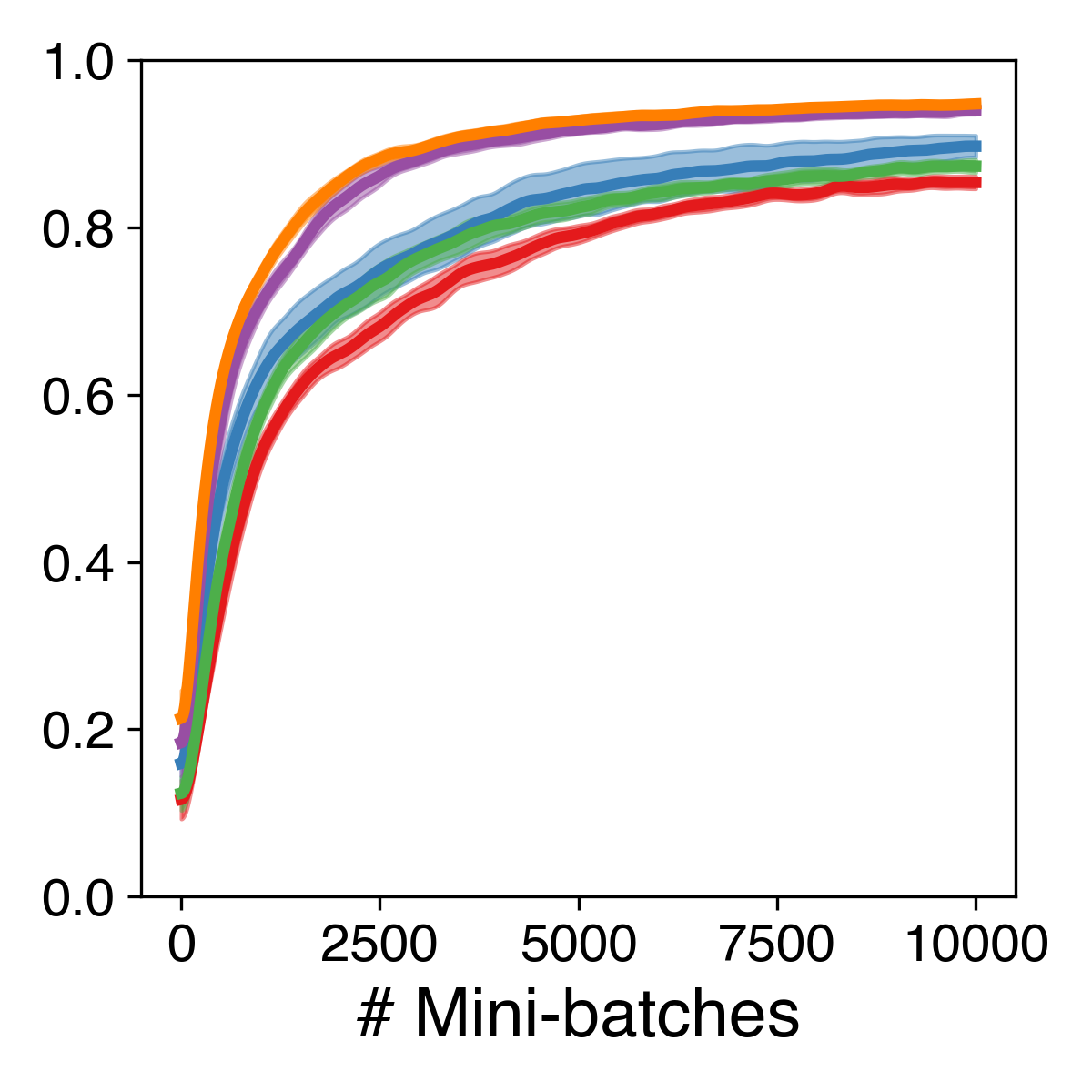}}
\caption{Mean validation accuracy across four seeds throughout offline training for (a) R-Randman, (b) T-Randman, and (c) SHD respectively. Shaded regions indicate one standard deviation.}
\label{fig:offline_acc}
\end{figure}

Figure~\ref{fig:offline_acc} compares OTPE and Approx OTPE against BPTT, OTTT, and OSTL for offline training. We compare performance across multiple datasets (R-Randman, T-Randman, SHD). We train a 2-hidden-layer model with a layer width of 128 for both variations of Randman. Performance on R-Randman is similar across methods (the range of their means is $1.3\%$), and both OSTL and OTTT beat BPTT, OTPE, and Approx-OTPE. However, when evaluated on T-Randman, OTPE and Approx-OTPE accuracy is on-par with BPTT (OTPE's smoothed validation accuracy on SHD is $\sim0.8\%$ lower, $\sim0.7\%$ higher on T-Randman, and $\sim0.1\%$ higher on R-Randman) while OSTL ($\sim7.4\%$ lower on SHD,  $\sim3.8\%$ lower on T-Randman, and $\sim1.1\%$ higher on R-Randman) and OTTT ( $\sim9.3\%$ lower on SHD, $\sim8.6\%$ lower on T-Randman, and $\sim0.7\%$ higher on R-Randman) underperform. All reported numbers are averages over multiple seeds, with SHD results also reported for the last 1000 minibatches. On SHD, Approx OTPE achieves $5.2\%$ lower validation accuracy than BPTT. This is $\sim{2.2}\%$ higher than OSTL's validation accuracy and $\sim{4}\%$ higher than OTTT. We provide training loss in \ref{ap:HP_search} and \ref{sec:add_plots}.

\begin{figure}[tbp]
\centering
\subfloat[]{
\includegraphics[width=.31\textwidth]{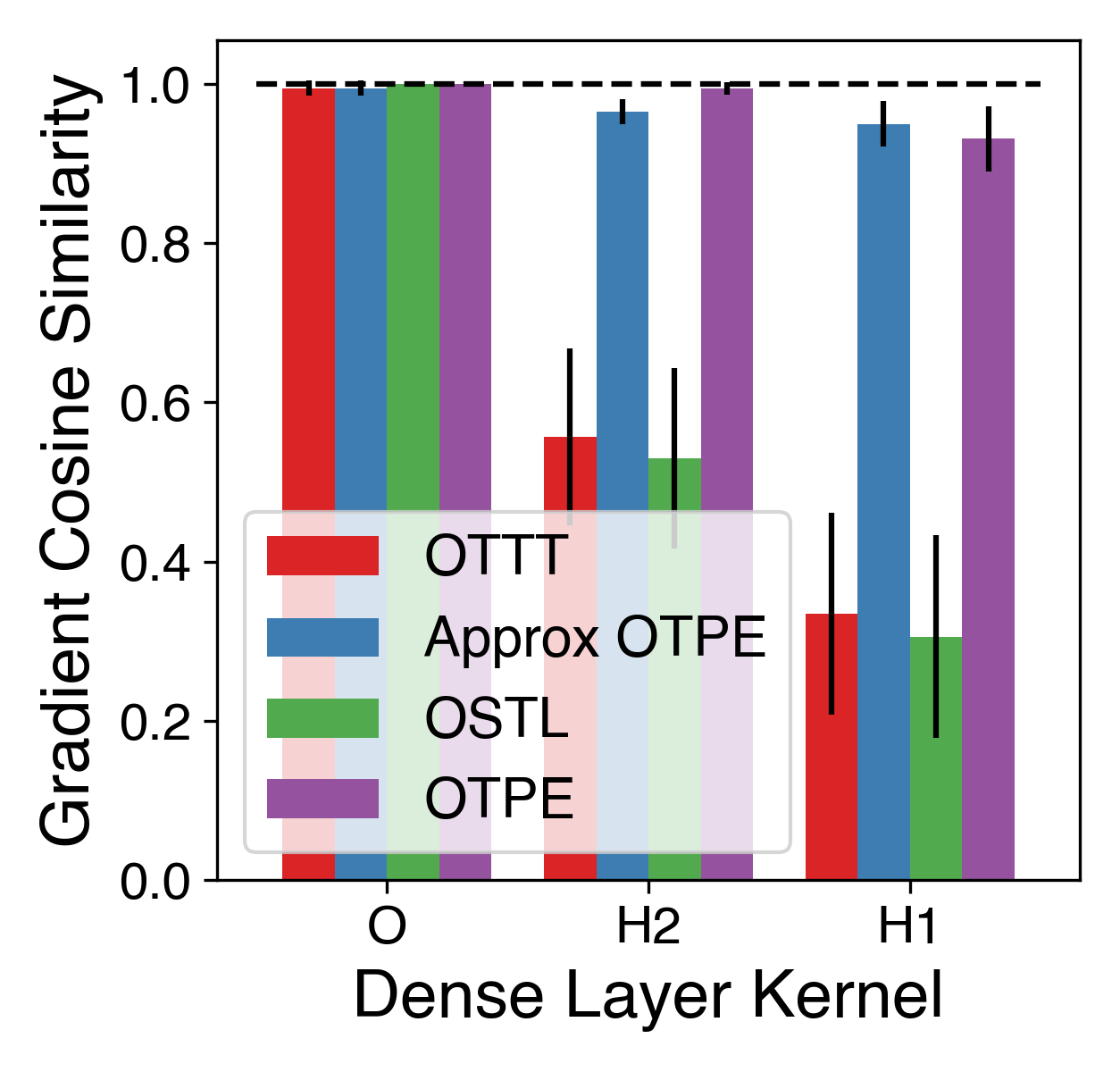}\hfill}
\subfloat[]{
\includegraphics[width=.3\textwidth]{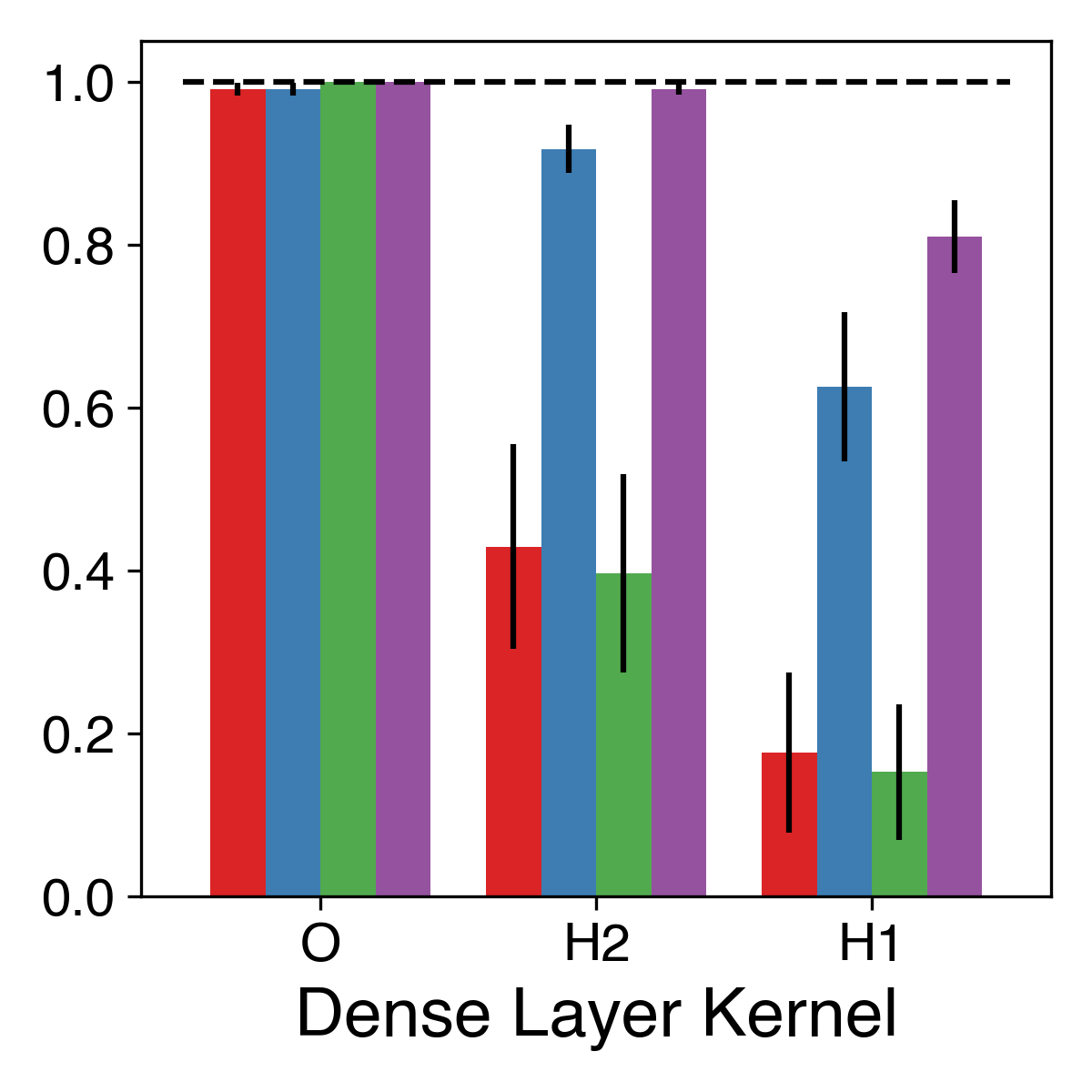}\hfill}
\subfloat[]{
\includegraphics[width=.3\textwidth]{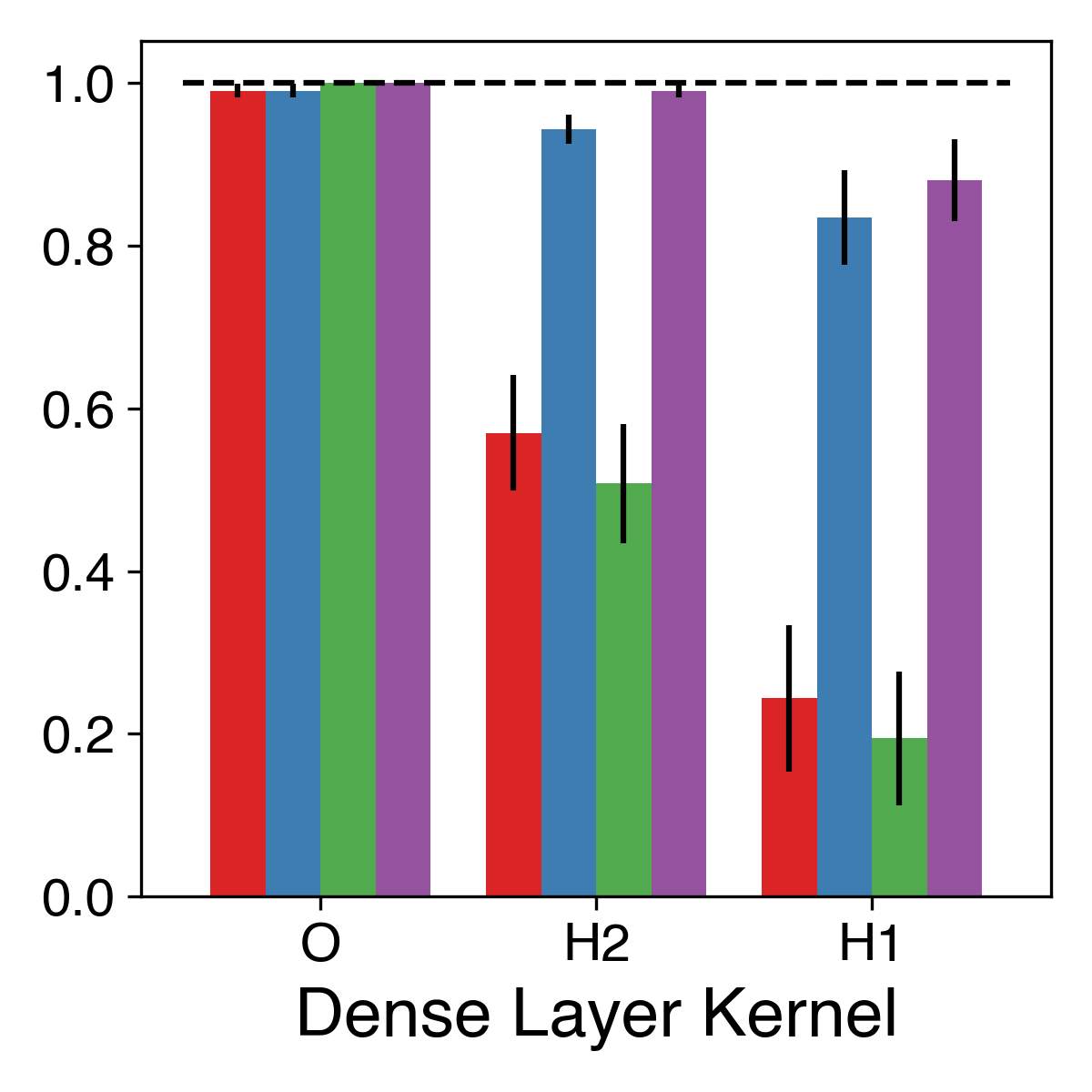}\hfill}
\caption{Mean gradient cosine similarity throughout training for each layer, evaluated on (a) R-Randman, (b) T-Randman, and (c) SHD. The output layer (O), has a higher cosine similarity than the earlier layers hidden-1 (H1) and -2 (H2) due to accumulating approximation error during backprop.}
\label{fig:lcos}
\end{figure}

Figure \ref{fig:lcos} shows layer-wise gradient cosine similarity between the gradient estimates of each algorithm (OTPE, Approx OTPE, OTTT, and OSTL), and BPTT for offline learning. We evaluate similarity across the three datasets for 2-hidden-layer networks with 128 layer width in both Randman configurations and a 2-hidden-layer network with 512 layer width. BPTT's gradients are identical to the gradients in the output layers of OSTL and OTPE due to their exact formulation. OTTT and Approximate OTPE are achieve an average $\ge0.99$  cosine similarity in the output layer. As expected, the gradient similarity with BPTT decreases for deeper layers across all algorithms. OSTL and OTTT incur a $\ge40\%$ reduction in alignment with BPTT in the second hidden layer (H2), which further decreases to $60\%$ in the first hidden layer (H1). However, OTPE and Approx OTPE are consistently better aligned with those generated by BPTT, with Approx OTPE above 0.6 in H1 for T-Randman compared to $\le0.2$ for OSTL and OTTT. We consistently observe these trends in layerwise gradient alignment across the evaluated datasets. Additionally, when observed over the training duration, we see consistent results for layerwise gradient-similarity. The output layers for OTTT and OTPE show improved gradient alignment with BPTT over time ( see Appendix \ref{sec:add_plots}). OTPE shows this increasing alignment with BPTT's gradient directions for the last hidden layer, whereas this trend reverses for OTTT. These trends are consistent across seeds, which may allow the slope of the trends to influence the standard deviation reported in Fig.~\ref{fig:lcos}. %While figure \ref{fig:lcos} reports mean values for each layer, there are trends in these values throughout training. The gradient similarity with OTTTT in the output layer increases throughout training, and this trend also exists for OTPE in the last hidden layer\ts{see appendix}.

% all methods achieve similar loss and validation accuracy, as seen in figures \ref{fig:offline_loss} and \ref{fig:offline_acc}. 
% In T-Randman, however, we see performance differences between the algorithms. The validation accuracy curves of OTPE and its approximation overlap with BPTT, while both OSTL and OTTT perform worse. Of the two, OSTL performs better. The training loss in T-Randman does not immediately drop as seen in R-Randman and SHD, although OSTL and OTTT are still the least performant. 

% OSTL and OTPE, which uses OSTL in its output layer, demonstrate 
\begin{figure}[tbp]
\centering
\subfloat[]{
\includegraphics[width=.31\textwidth]{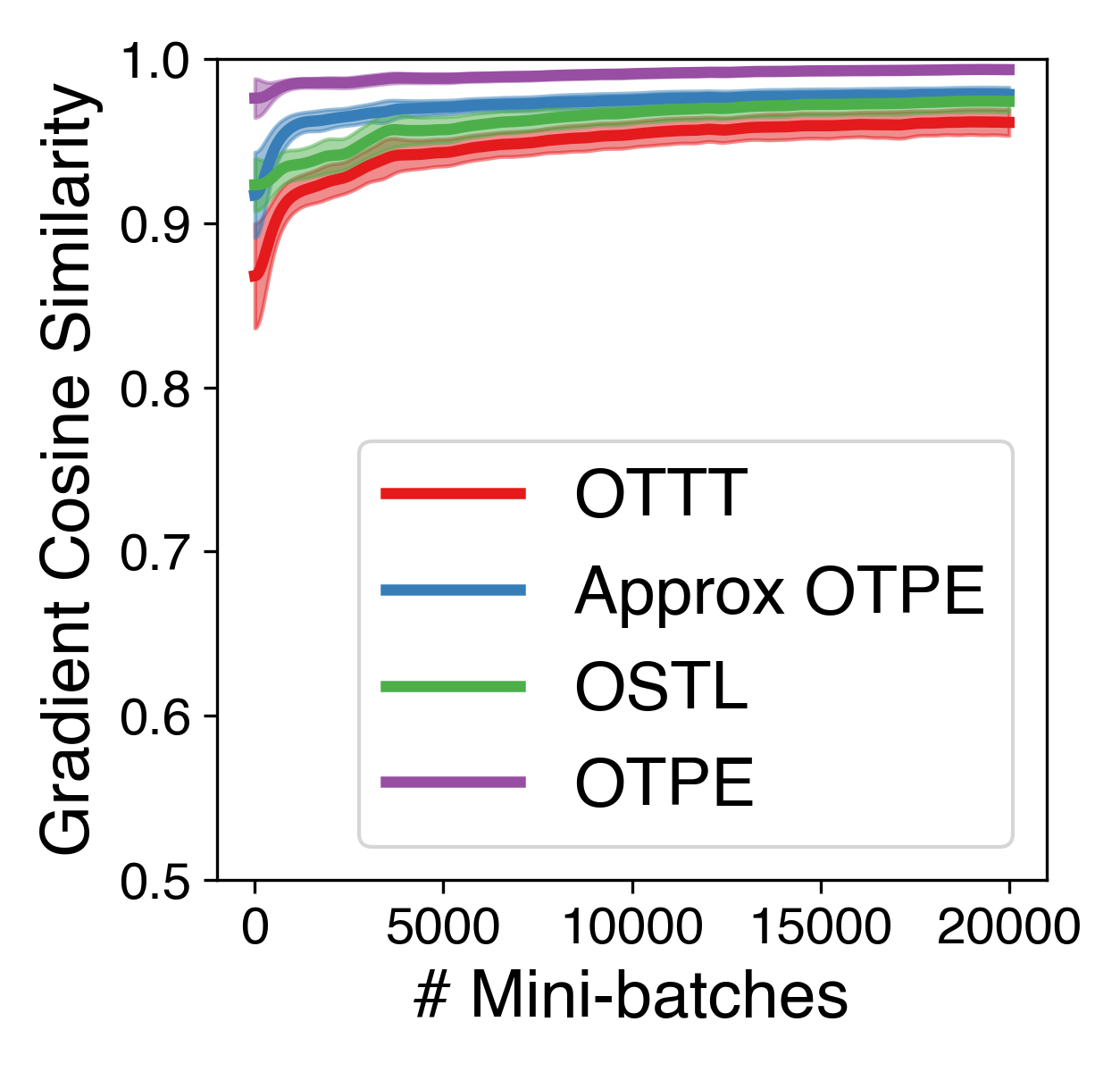}\hfill}
\subfloat[]{
\includegraphics[width=.30\textwidth]{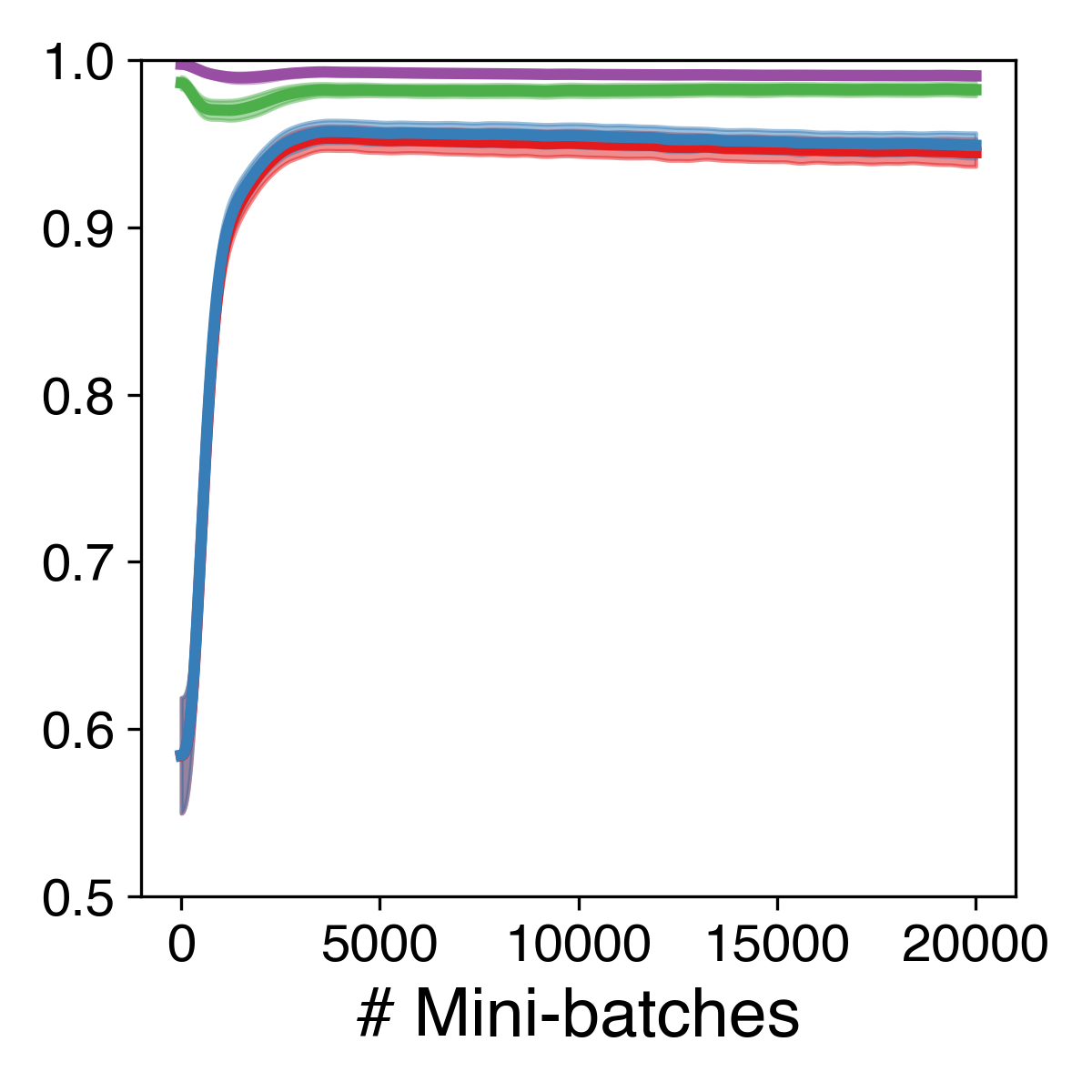}\hfill}
\subfloat[]{
\includegraphics[width=.30\textwidth]{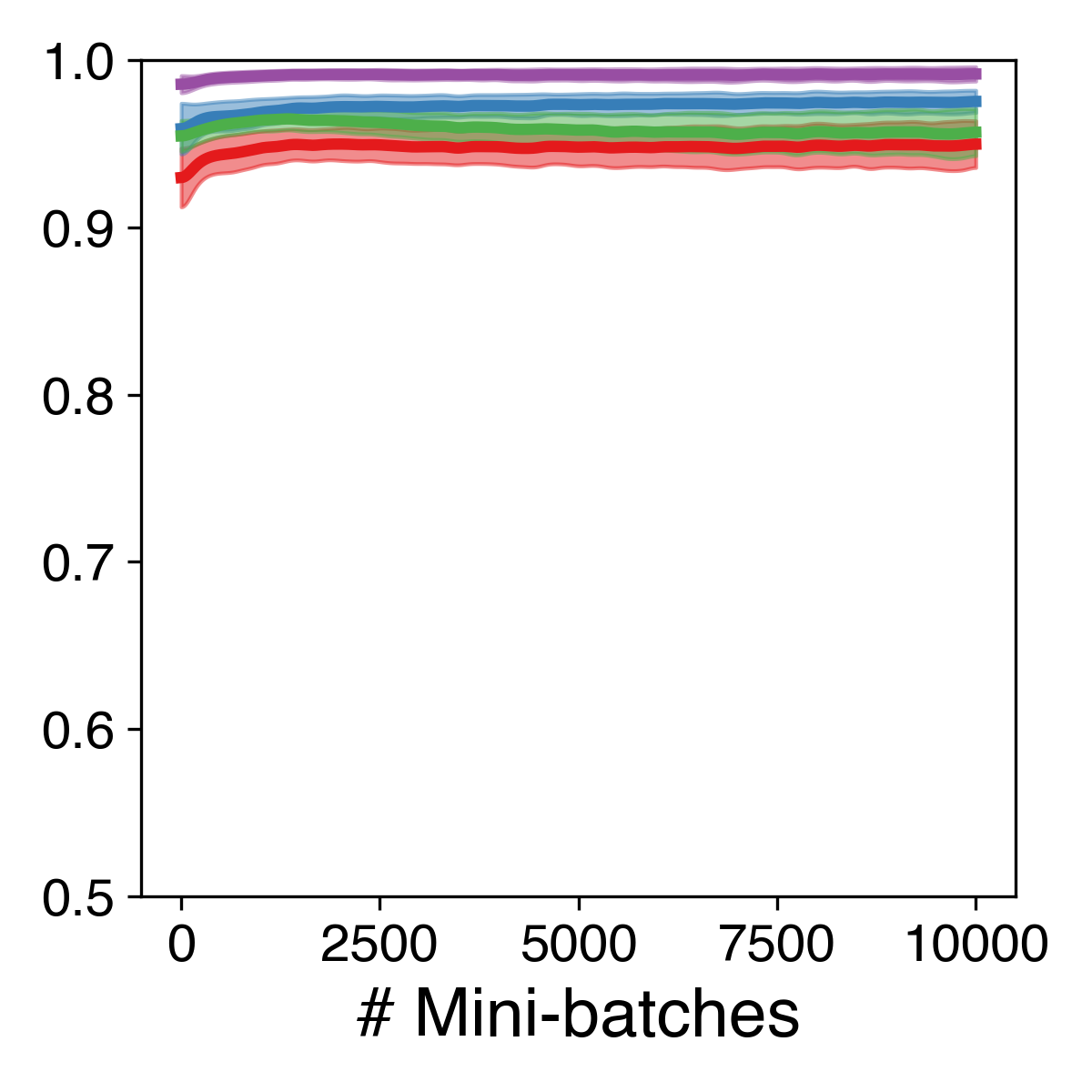}}
\caption{Model-wide, mean gradient cosine similarity over training duration for (a) R-Randman, (b) T-Randman, and (c) SHD. Shaded regions indicate one standard deviation.}
\label{fig:mcos}
\end{figure}

As shown in Fig~\ref{fig:mcos}, OTPE consistently achieves the highest model-wise gradient cosine similarity with BPTT over the training duration. When evaluated on T-Randman, OTTT and Approximate OTPE are less aligned with BPTT early in the training ($0.6$ initially) before stabilizing above $0.9$. For SHD and R-Randman, OTPE and its approximation consistently achieve a higher alignment (average of $0.99$ and $0.97$, respectively) than OSTL and OTTT (average of $0.96$ and $0.95$) .

% one chooses a center point θ ∗ in the graph, and chooses two direction vectors,
%  and computing the loss values  along the two most significant directions of variance in weight space for the models

% I that's probably enough for the loss landsc ape explanation?
% orthogonalitz is ofc something we would have to caclulate, but we can for sure say they end up with a substiantially different model -> but i guess in this high dimensional space orthogonality is somewhat guranteed? (when they are so far away)
% agreed, I want a good way of saying that the final configuration for OSTL and OTTT are orthogonal in wieght spce to teh one trained using BPTT or at least they appear to be so? 
% good point. Yeah, then we can leave the quantifiable part to later and just talk about it qualitatively for now. Thanks!
% that's a good strategy

\begin{figure}[tbp]
\centering
\subfloat{
\includegraphics[width=.45\textwidth]{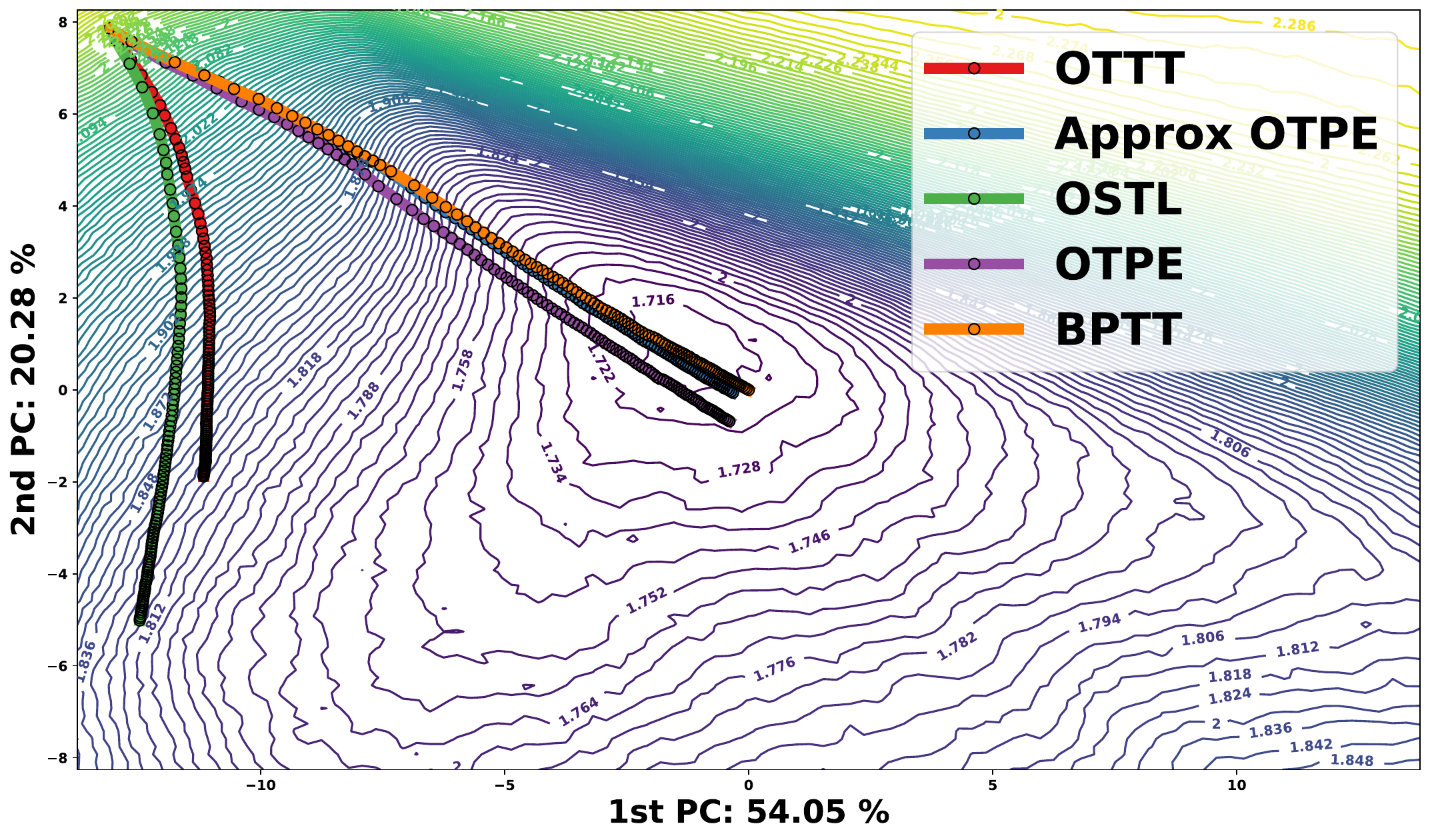}\hfill}
\subfloat{
\includegraphics[width=.45\textwidth]{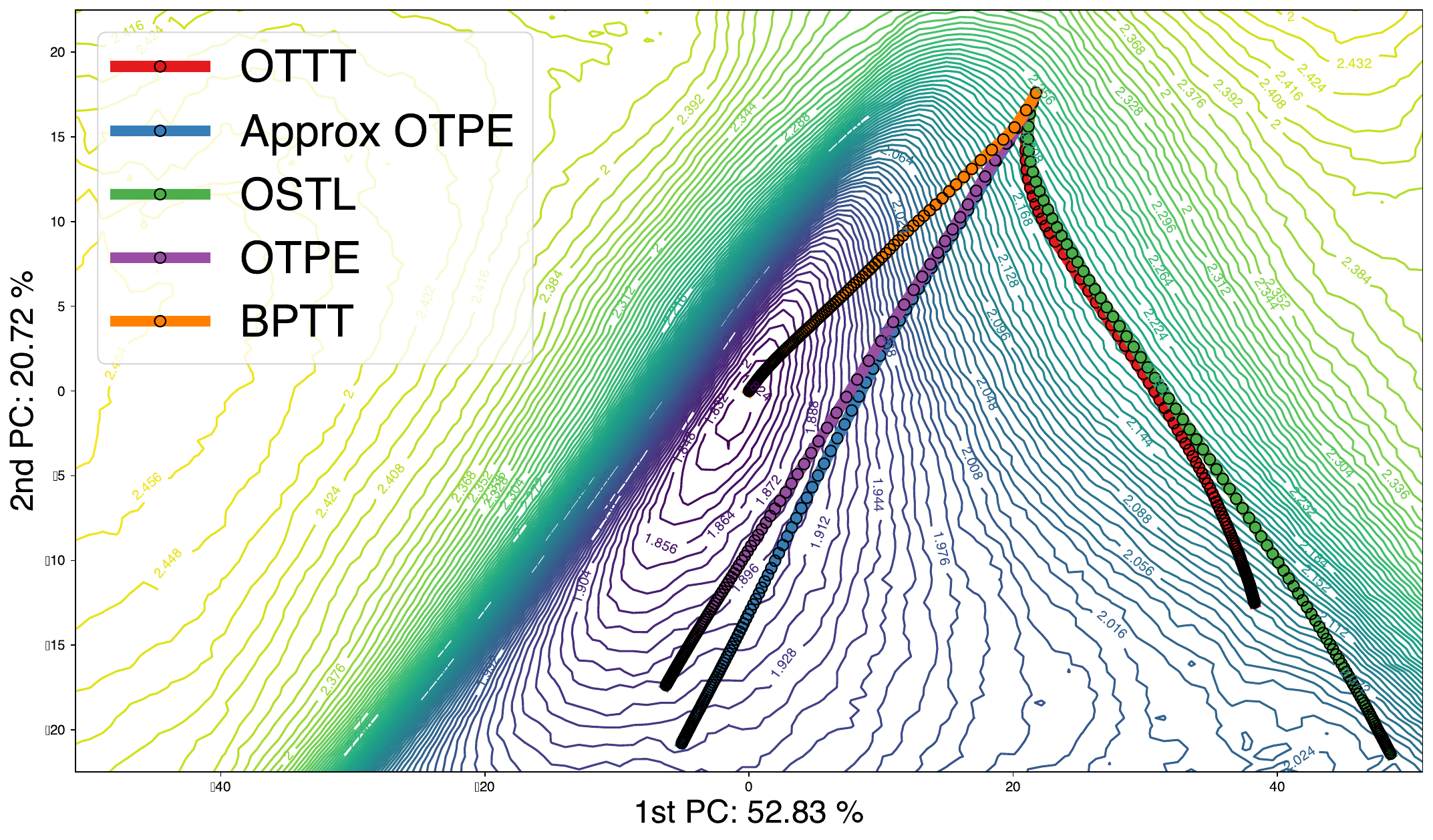}\hfill}
\caption{Evaluations of offline learning through the loss landscapes of the different algorithms for (a) R-Randman and (b) T-Randman, evaluated over the validation set. Evaluations are conducted every 200 mini-batches of training with BPTT's model. We observe high similarity between BPTT's model and those trained by OTPE and Approx OTPE in the model weight-space. The decreasing space between model evaluations indicates model parameter convergence as training progresses.}
\label{fig:loss_landscape}
\end{figure}

% shows how the optimization paths between the approximations and BPTT diverge and their success in minimizing the loss. OTPE and its approximation closely follow the same training trajectory as BPTT in R-Randman. OSTL and OTTT, on the other hand, diverge from BPTT's path early in training despite reaching a similar loss. In T-Randman, BPTT reaches the best local minimum, followed by OTPE and its approximation. While not as aligned as in R-Randman, OTPE and its approximation again appear to optimize more similarly to BPTT than OSTL and OTTT.
We visualize the training loss landscape for different algorithms on the R-Randman and T-Randman tasks (see Figure~\ref{fig:loss_landscape} (a) and (b)), as described in~\cite{li2018visualizing}. The loss contours are determined by picking a centre point ($\theta*$) and two direction vectors (the initial ($\delta$) and final ($\nu$) BPTT model) and then plotting $f(\alpha, \beta) = L(\theta* + \alpha \delta + \beta \nu)$. The trajectories are rendered by mapping a model, every 200 minibatches during training, into the loss landscape. The strong alignment between BPTT, OTPE, and Approx. OTPE for R-Randman indicates a remarkable similarity between the trained models. For T-Randman, the models trained using OTPE and Approx. OTPE remain more closely aligned to BPTT compared to those trained using OSTL and OTTT. Across these different datasets, the training trajectories of OSTL and OTTT diverge the most from the other models, indicating a significantly altered trained model configuration compared to BPTT. 

\begin{figure}[]
\centering
\subfloat[]{
\includegraphics[width=.31\textwidth]{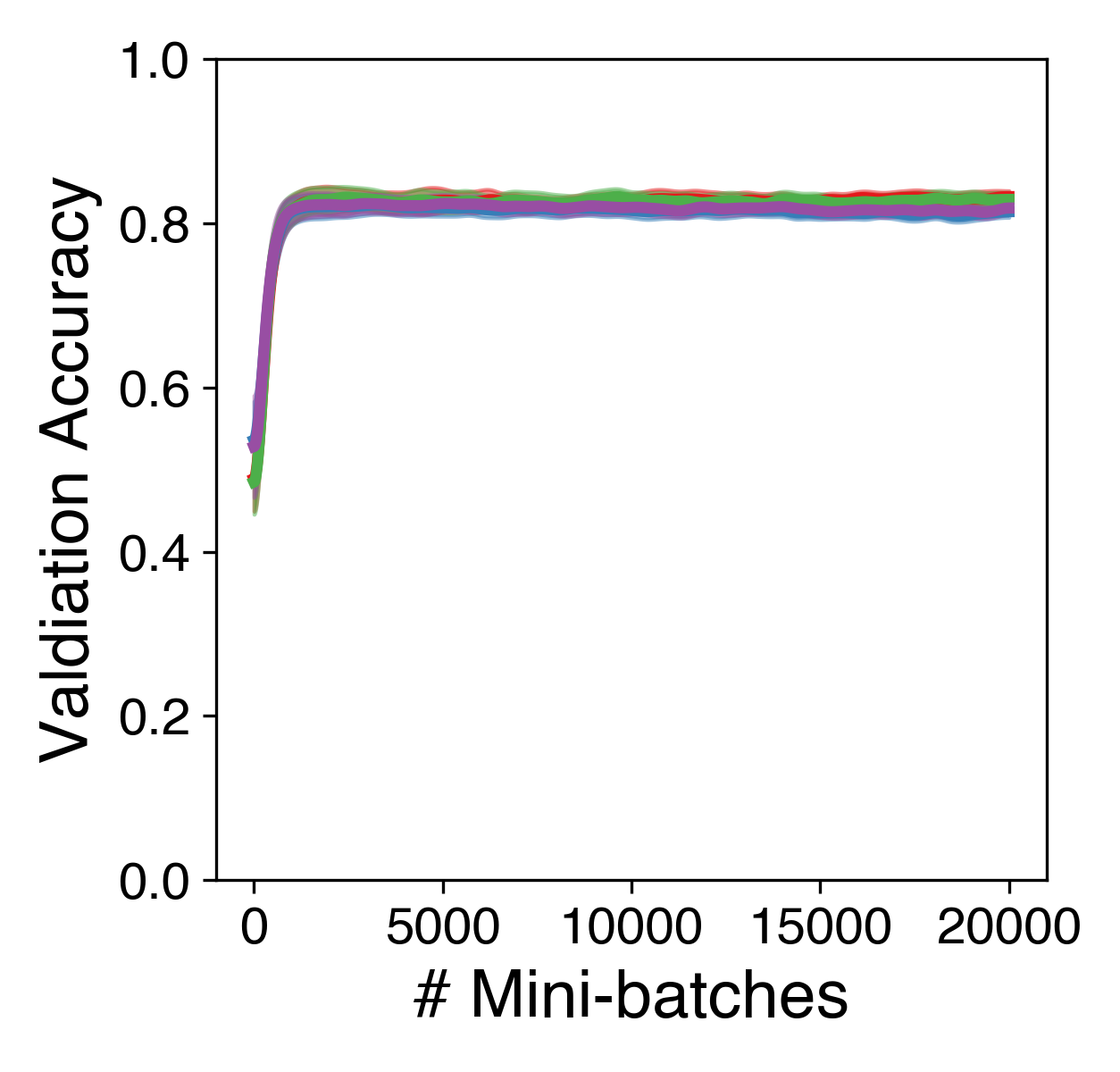}\hfill}
\subfloat[]{
\includegraphics[width=.30\textwidth]{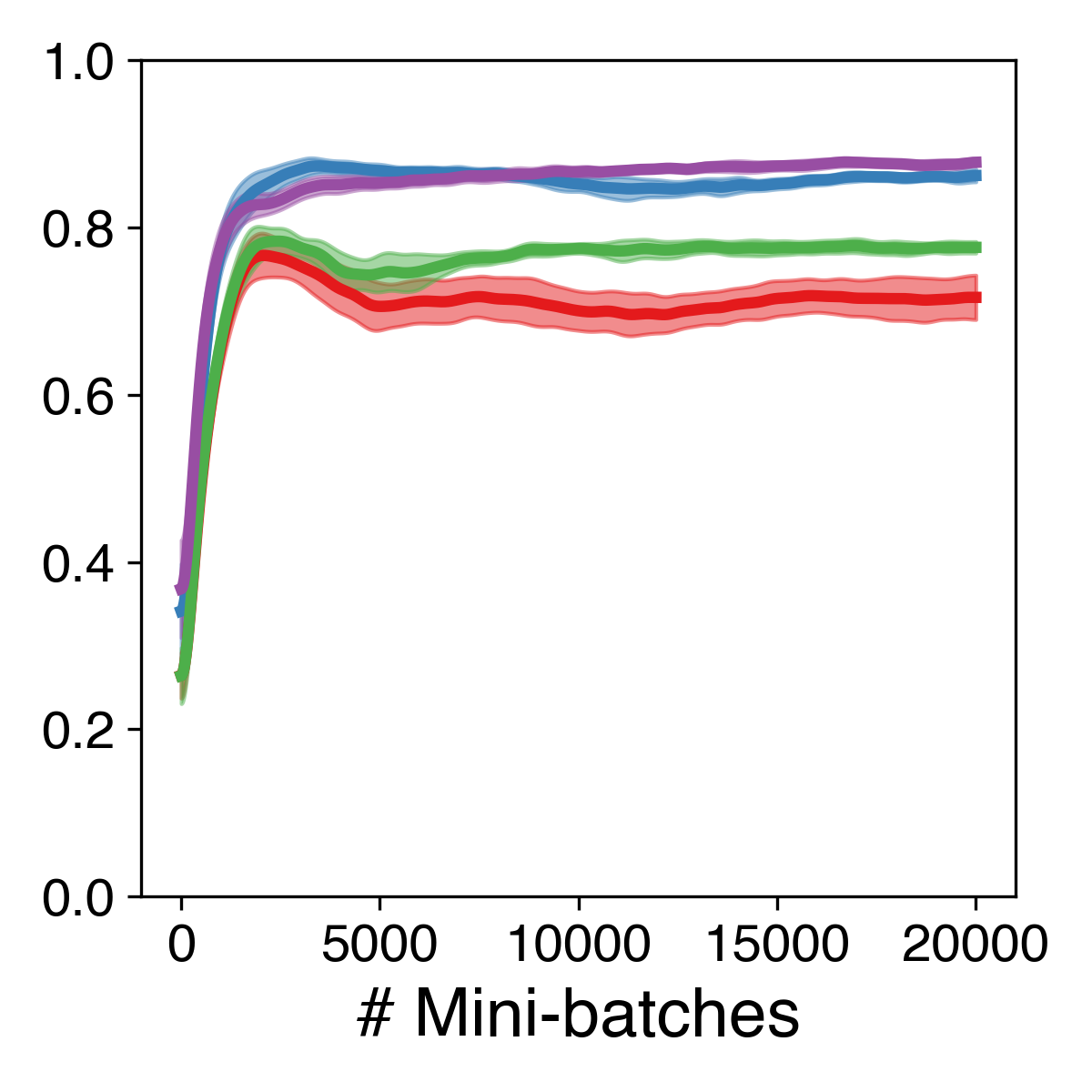}\hfill}
\subfloat[]{
\includegraphics[width=.30\textwidth]{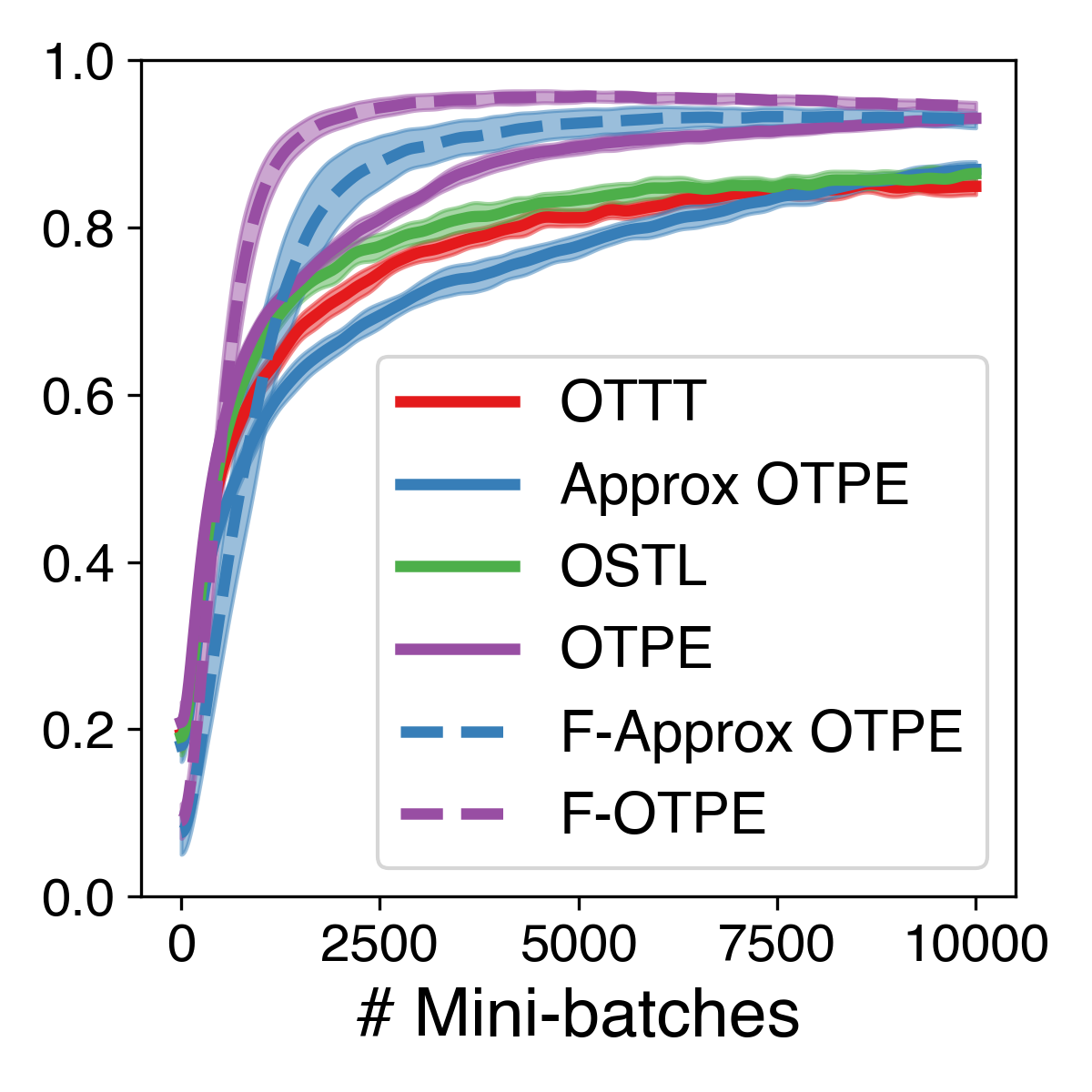}}
\caption{Mean validation accuracy across four seeds for online training evaluated on (a) R-Randman, (b) T-Randman, and (c) SHD. Shaded regions indicate one standard deviation across seeds.}
\label{fig:online_acc}
\end{figure}

Table~\ref{tab:shdtab}, summarizes our performance results for SHD evaluated with different model configurations. We found that OTPE and its approximation perform best with five layers while OTTT, OSTL, and BPTT perform best with three layers. The performance difference between the three-layer and five-layer models is within one standard deviation for BPTT. In contrast OTTT and OSTL incur a substantial drop of $5\%$ and $3.4\%$ in test accuracy after online training, respectively, when training a five-layer model. BPTT unsurprisingly dominates test performance on SHD, but OTPE consistently outperforms OTTT and OSTL. We additionally observe high performance for OTPE across different model and training hyperparameters (data in Appendix \ref{ap:HP_search}). This finding is consistent with \cite{bauer2023exodus}, where the approximate gradient calculation of \cite{slayer} is more sensitive to the surrogate derivative slope than exact gradient calculation.

% are for our algorithm are 
Results comparing online learning performance across the different algorithms are shown in Fig.~\ref{fig:online_acc}. Online training performance on R-Randman is consistent with those observed for offline training, with all approaches delivering similar accuracy (see training loss in Appendix \ref{sec:add_plots}). The performance difference across OTPE, its approximations, OSTL, and OTTT is more apparent for T-Randman and SHD. Although we observed higher test accuracy for online learning on SHD than for offline learning, we attribute this to our hyperparameter search (see Appendix \ref{ap:HP_search}). As seen in Fig.~\ref{fig:online_acc} (c), when evaluated on SHD, we observe that both F-OTPE and F-Approx OTPE converge in performance earlier than OTPE. F-OTPE reaches its peak average performance on SHD after 4,500 mini-batches of online training. Meanwhile, OTPE does not appear to have reached peak validation accuracy even after training on 10,000 mini-batches. Additional results for online training with 5 layers and 512 layer width on SHD are provided in Appendix \ref{sec:add_plots}.

\begin{table}

  \centering
  \begin{tabular}{lcccc}
    \cmidrule(r){1-5}
    Name     & Depth & Width  & Offline Acc $\pm\sigma$& Online Acc $\pm\sigma$ \\
    \midrule

    OTTT & 3 & 128 & $64.3 \% \pm 1.0$& $66.7 \% \pm 1.1$ \\
    OSTL& 3 & 128 & $66.7 \% \pm 1.0$& $68.5 \% \pm 1.9$ \\
    BPTT& 3 & 128 & $\mathbf{73.9} \% \pm 2.5$& N/A \\
    OTPE & 3 & 128 & $72.5 \% \pm 0.7$& $\mathbf{73.6} \% \pm 1.4$\\
    A. OTPE& 3 & 128 & $67.0 \% \pm 1.0$& $69.8 \% \pm 1.2$ \\
 F-OTPE& 3& 128& N/A&$73.3 \% \pm 0.9$ \\
 F-A. OTPE& 3& 128& N/A&$71.2 \% \pm 1.7$ \\

    \cmidrule(r){1-5}

    OTTT & 3 & 512& $70.5 \% \pm 1.5$& $71.2 \% \pm 0.8$ \\
    OSTL& 3 & 512& $70.5 \% \pm 1.5$& $70.6 \% \pm 0.7$ \\
    BPTT& 3 & 512& $\mathbf{78.1} \% \pm 1.0$& N/A \\
    OTPE & 3 & 512& $75.2 \% \pm 0.5$& $\mathbf{75.4} \% \pm 0.5$\\
    A. OTPE& 3 & 512& $71.2 \% \pm 1.1$& $71.8 \% \pm 1.1$ \\
    F-OTPE& 3& 512& N/A& $75.3 \% \pm 0.3$ \\

    F-A. OTPE& 3& 512& N/A& $71.5 \% \pm 1.2$ \\

    \cmidrule(r){1-5}

    OTTT & 5& 512& $63.5 \% \pm 2.1$& $66.2 \% \pm 0.7$ \\
    OSTL& 5& 512& $63.9 \% \pm 1.0$& $67.2 \% \pm 1.0$ \\
    BPTT& 5& 512& $\mathbf{77.7} \% \pm 0.6$& N/A \\
    OTPE & 5& 512& $76.4 \% \pm 1.2$& $\mathbf{76.7} \% \pm 0.7$\\
    A. OTPE& 5& 512& $74.3 \% \pm 0.9$& $\mathbf{76.7} \% \pm 0.8$\\

    F-OTPE& 5& 512& N/A& $74.1 \% \pm 1.0$ \\
    F-A. OTPE& 5& 512& N/A& $72.6 \% \pm 1.3$ \\

    \bottomrule
  \end{tabular}
\caption{Summary of test accuracy on SHD for different model configurations. Results for online learning report highest accuracy determined after learning rate search (Appendix \ref{ap:HP_search}).}
\label{tab:shdtab}
\end{table}

\section{Discussion and Conclusions}
We propose OTPE and its approximations to facilitate efficient online training in SNNs, by capturing temporal effects typically omitted by similar algorithms. While maintaining similar scalability to OSTL in both compute and memory costs, OTPE produces superior results while remaining layer-local. OTPE and its approximation demonstrate greater alignment to exact gradients in the hidden layers, which may be more beneficial in tasks requiring greater network depth. The training trajectories in the loss landscape consistently demonstrate that models trained with OTPE or its optimizations are closer in model weight-space to BPTT models than those trained using OTTT and OSTL. We evaluated our algorithms on SHD and on rate- and temporal- variations of the Randman task (R-Randman and T-Randman). We observe similar performance across all algorithms when evaluated on R-Randman. However, the temporal influence approximations used by OSTL and OTTT result in degraded performance on temporal tasks like T-Randman and SHD, with OTPE and its variants consistently outperforming them across model configurations and datasets.

\section{Reproducibility}

The code for our experiments is at https://github.com/Intelligent-Microsystems-Lab/OTPE. The \texttt{readme} file outlines the scripts for generating data and plots.

% \section*{References}

% \medskip

\small

\bibliography{main}

\newpage
\appendix
\section{Appendix}
\subsection{Randman and SHD hyperparameters}\label{ap:hyperparam1}

Our main parameters are listed here. All others are accessible in the code.

Our Randman parameters are as follows (see \verb|generate_randman_data.py| and \verb|randman_dataset.py| in code):
\begin{itemize}
    \item 3-dimensional manifolds
    \item 10 manifolds to classify
    \item alpha = 1.
    \item neurons = 50
    \item time-steps = 50
    \item batch size = 128
    \item Each batch is randomly sampled from the underlying manifolds
\end{itemize}

Our SHD parameters are as follows (see \verb|generate_SHD_data.py| in code):
\begin{itemize}
    \item time-steps = 50 (reduced to 50 by binning)
    \item neurons = 700
    \item $10\%$ of the training set is used for validation. The reported test accuracy uses the model parameters with the best validation accuracy, which is evaluated after each training batch.
\end{itemize}

\subsection{SHD hyperparameter sweeps}\label{ap:HP_search}

\begin{figure}[!htp]
\centering
\includegraphics[width=\textwidth]{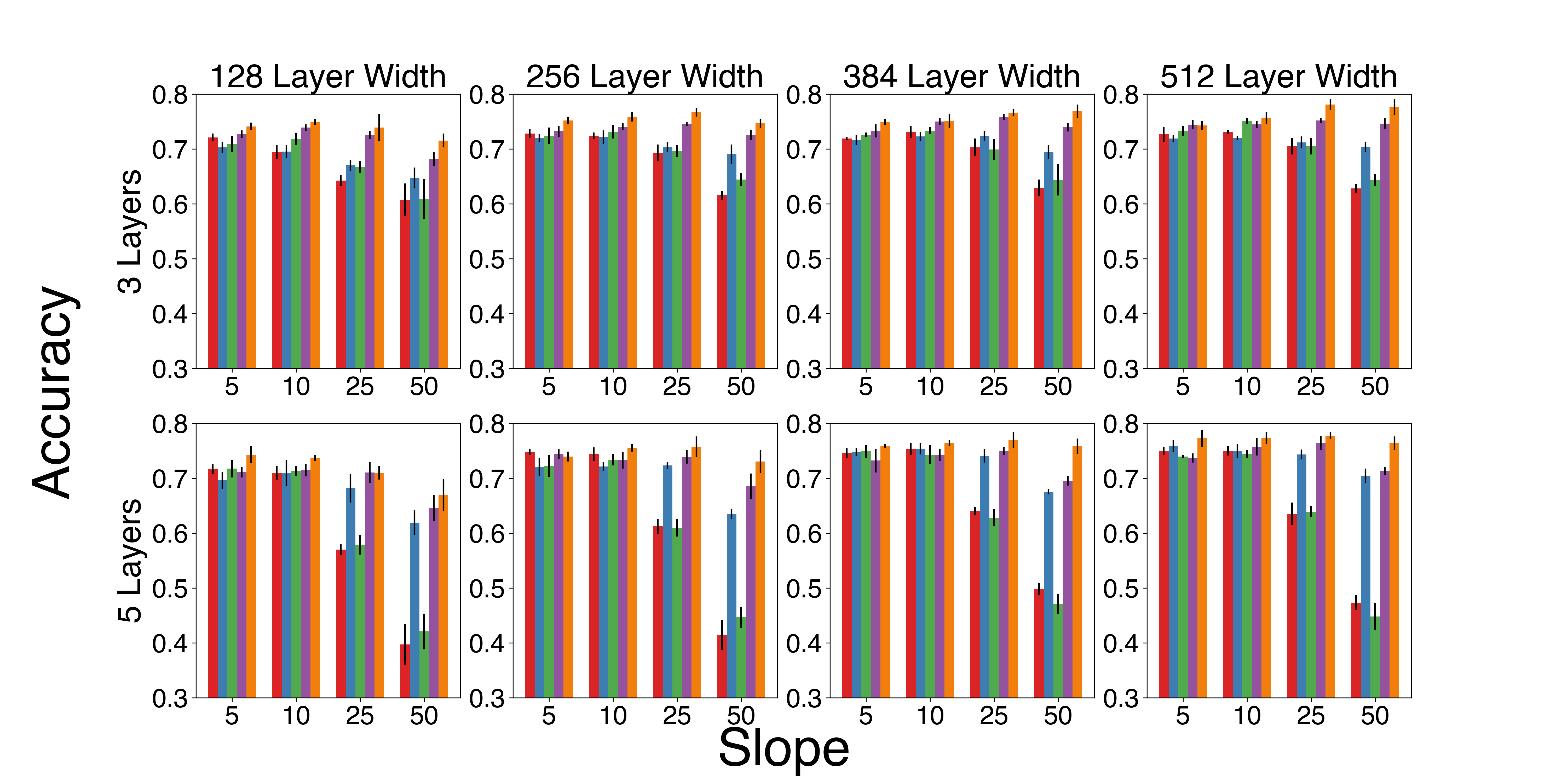}
\caption{Accuracy results during hyperparameter search.}\label{fig:hpsearchacc}
\end{figure}

We evaluate offline training on SHD across multiple model sizes and fast sigmoid slopes. One goal of this hyperparameter search is to determine which slope and model size yield the best performance. Another is to identify any trends across model size and slope. As seen in Fig.~\ref{fig:hpsearchacc}, BPTT consistently performs best. With the exception of 128 layer width, BPTT achieves the highest average score with a slope of 25. Interestingly, we see that going from three to five layers increases the performance of shallower slopes. Using five layers, however, appears to result in poorer performance from OTTT and OSTL, though BPTT also marginally drops. OTPE and its approximation appear to benefit from increased layer depth. These results indicate OTPE has better depth scalability than OTTT and OSTL, especially at sharper slopes.

When we only look at the offline performance of the approximate algorithms, we see OTPE achieving the best average accuracy in the entire sweep and does so using the best slope for BPTT. OTPE also achieves the highest average accuracy in most configurations while also having the most consistent performance across all configurations.

\begin{figure}[H]%[htp]

\centering
\subfloat[]{
\includegraphics[width=.5\textwidth]{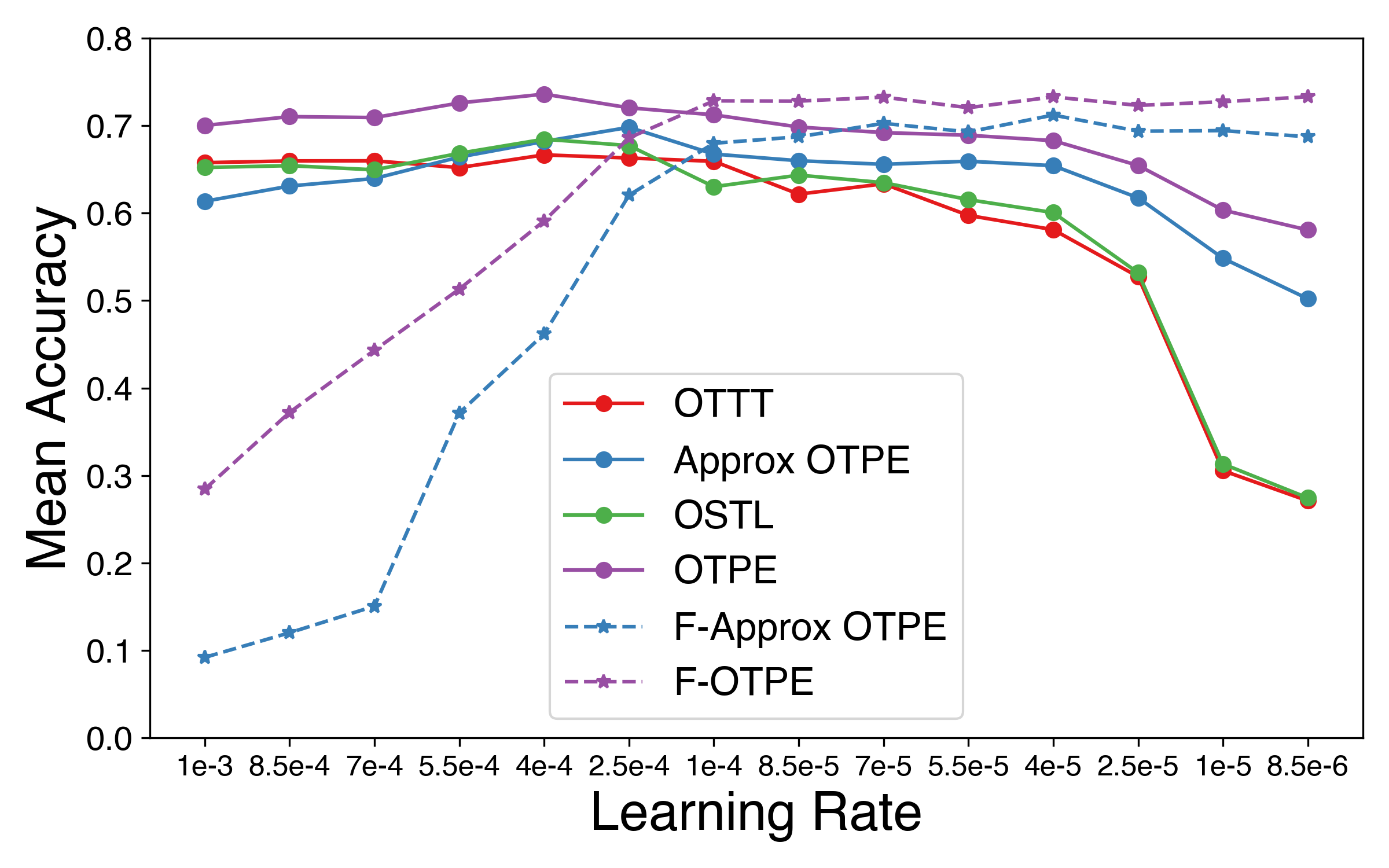}}
\subfloat[]{
\includegraphics[width=.5\textwidth]{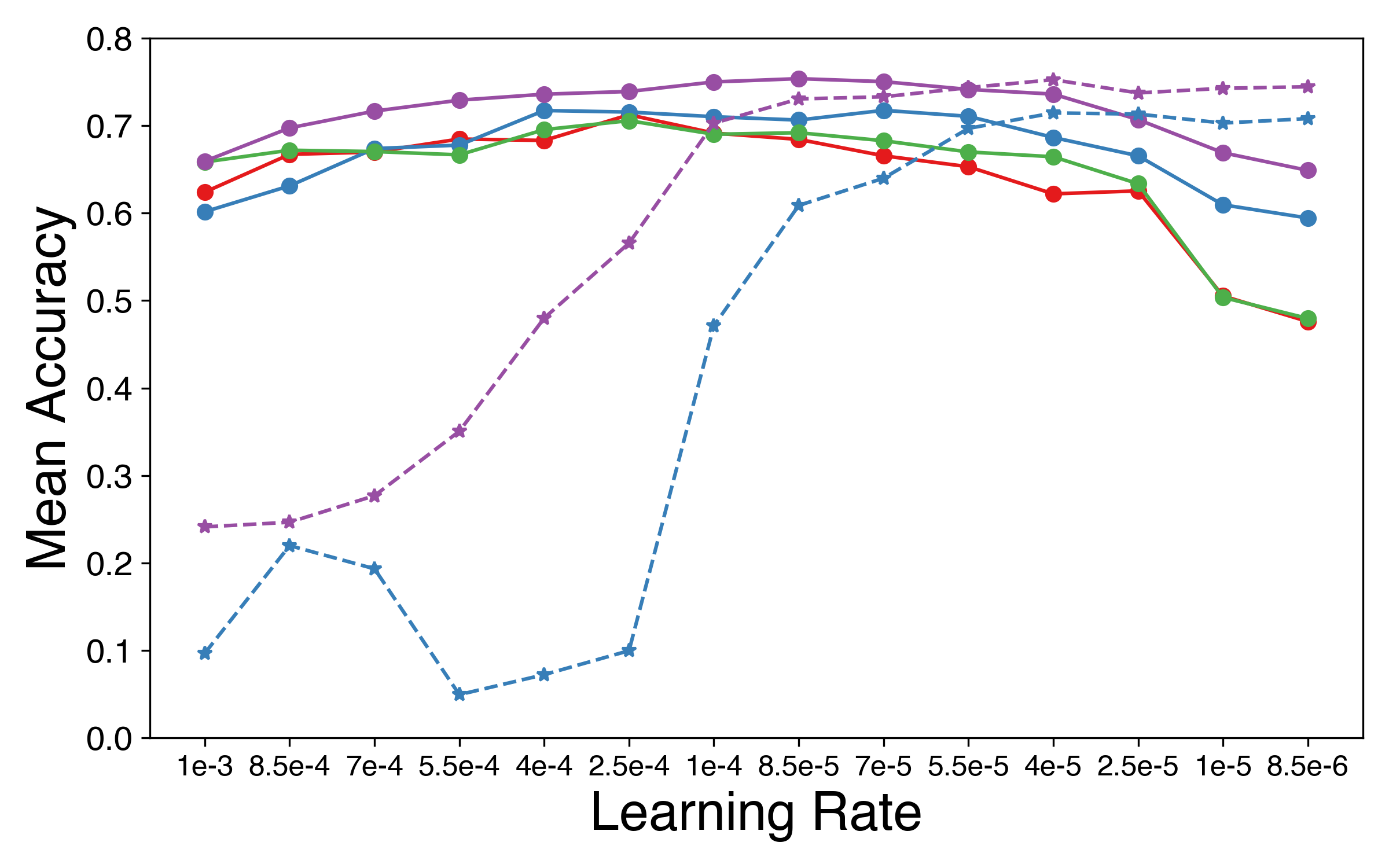}}\\
\subfloat[]{
\includegraphics[width=.5\textwidth]{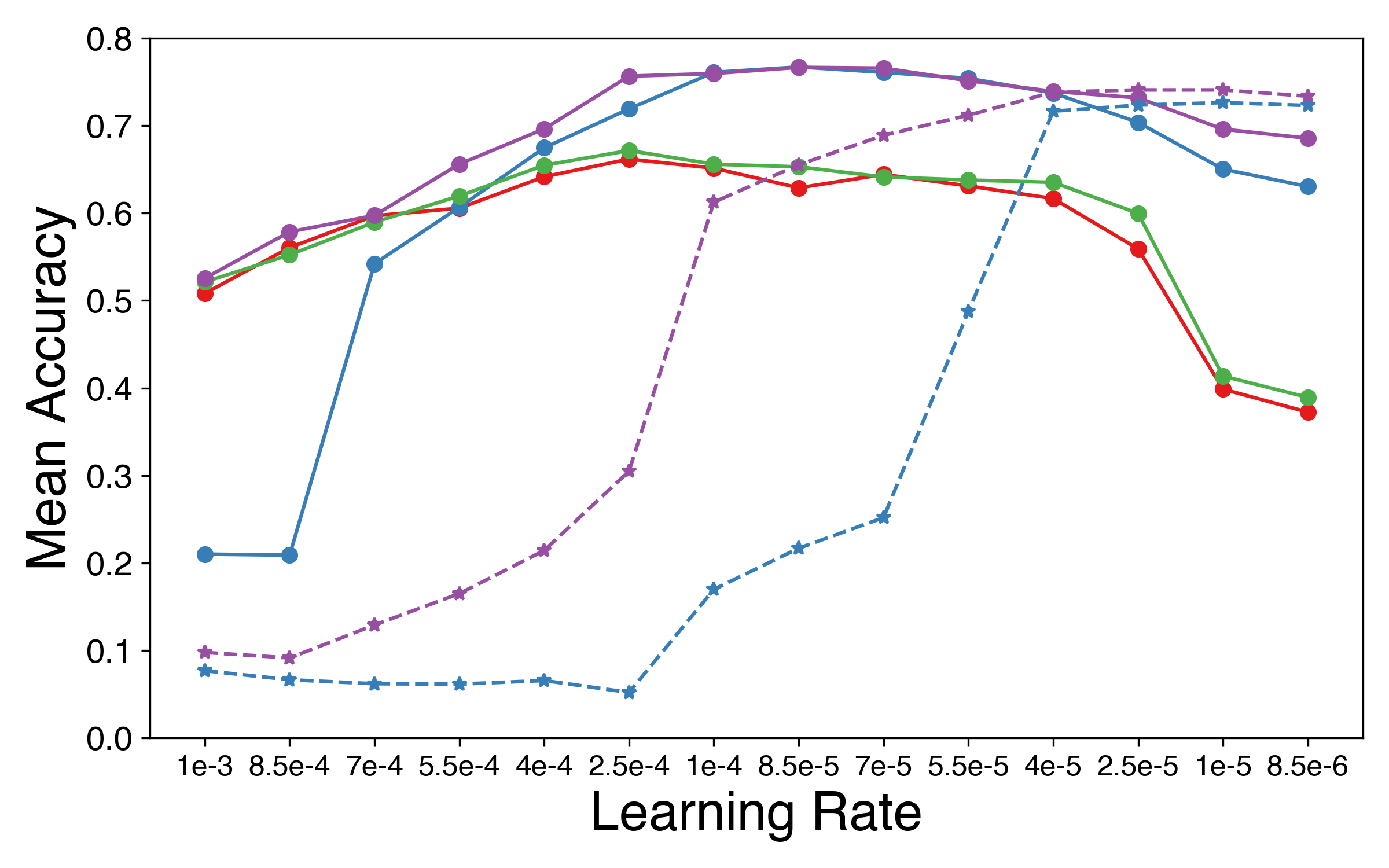}}

\caption{Learning rate hyperparameter search for online learning on SHD. Each point is the mean accuracy across 4 seeds. Figures (a), (b), and (c) are model configurations 3 layers with 128 layer width, 3 layers with 512 layer width, and 5 layers with 512 layer width, respectively.}
\label{fig:lr_search}

\end{figure}

\subsection{Additional training plots}\label{sec:add_plots}

\begin{figure}[!htp]

\centering
\subfloat[]{
\includegraphics[width=.34\textwidth]{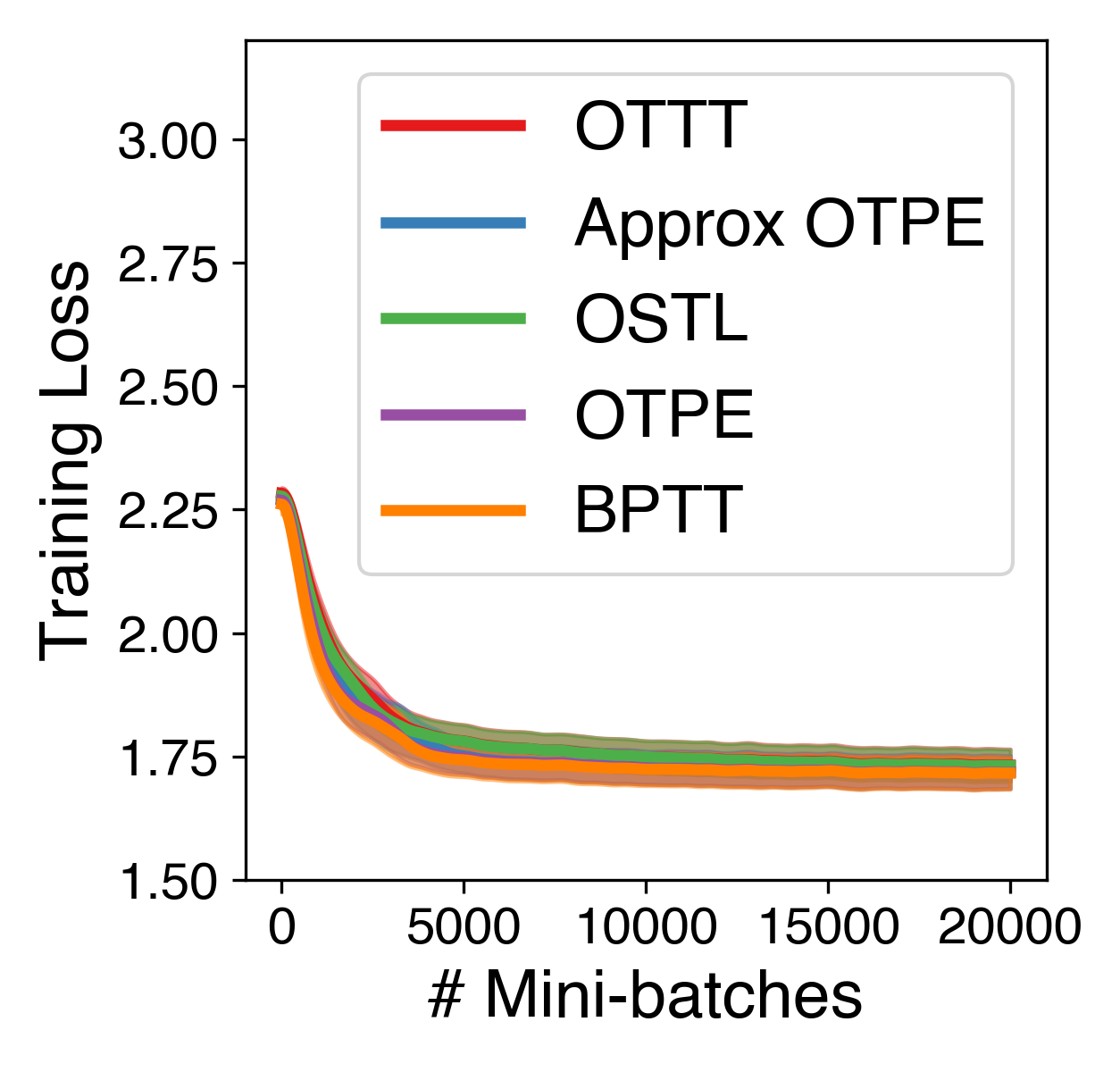}\hfill}
\subfloat[]{
\includegraphics[width=.33\textwidth]{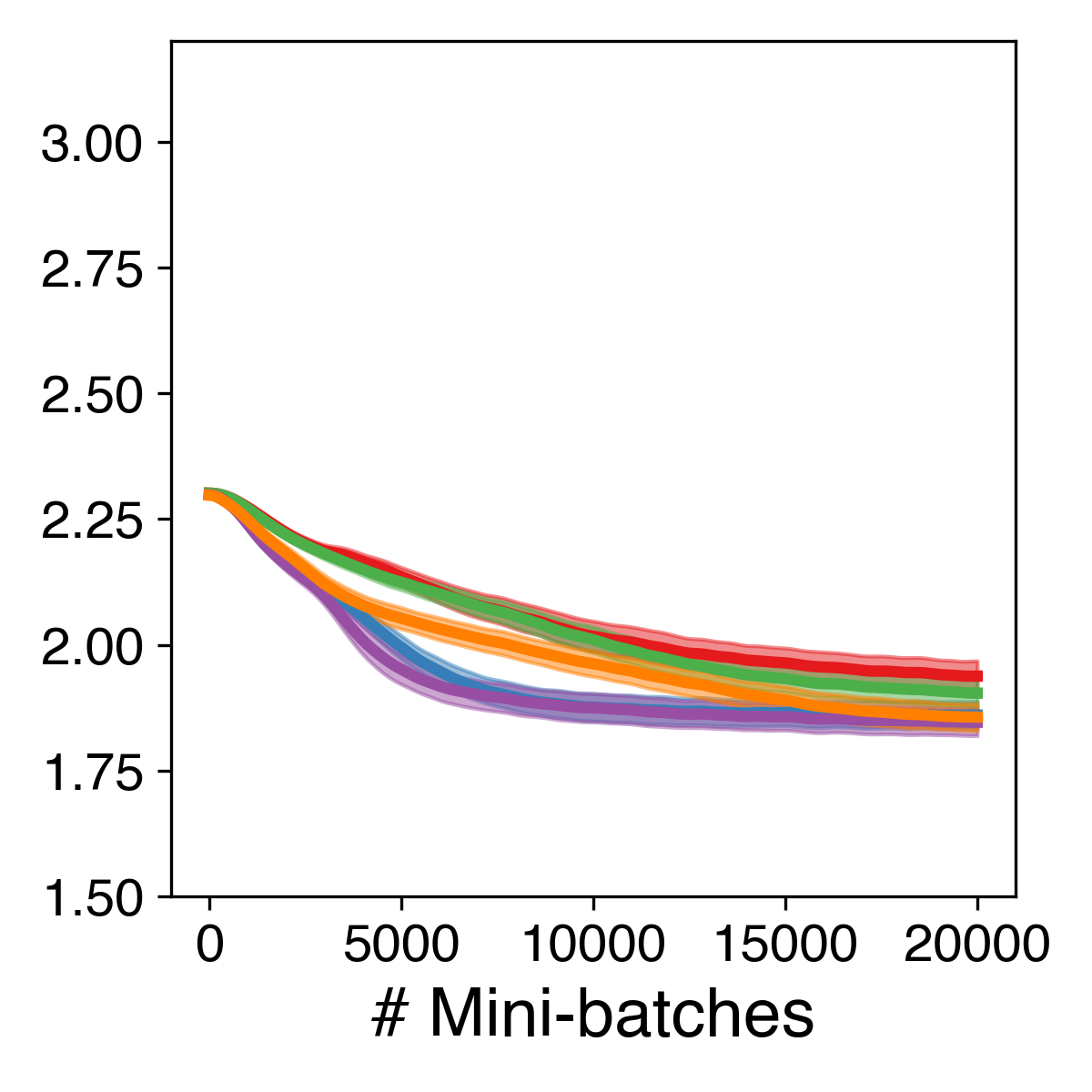}\hfill}
\subfloat[]{
\includegraphics[width=.33\textwidth]{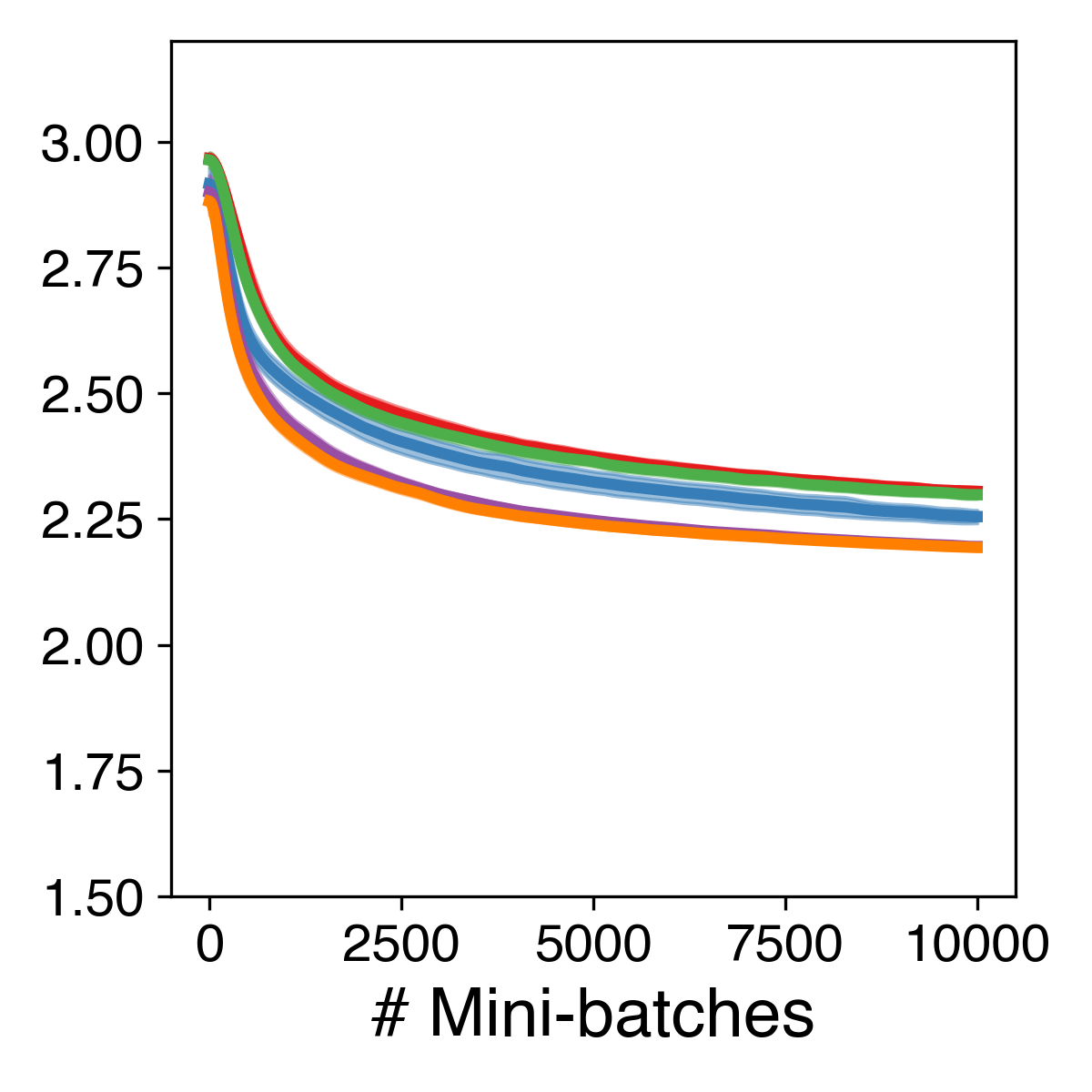}}
\caption{Mean training loss across four seeds throughout offline training. Figures (a), (b), and (c) are R-Randman, T-Randman, and SHD, respectively. Shaded regions are one standard deviation.}
\label{fig:offline_loss}
\end{figure}

\begin{figure}[!htp]

\centering
\subfloat[]{
\includegraphics[width=.34\textwidth]{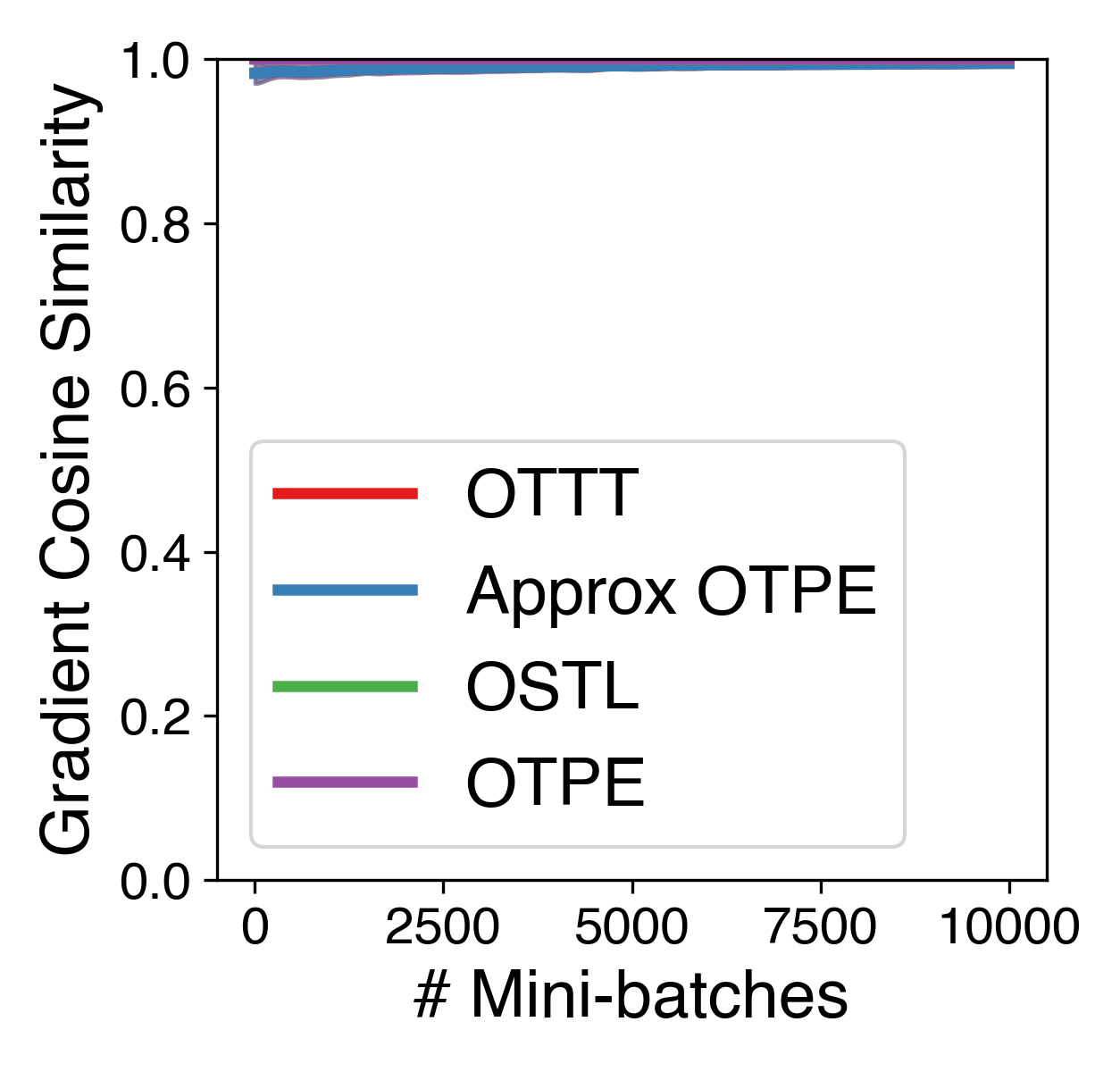}\hfill}
\subfloat[]{
\includegraphics[width=.33\textwidth]{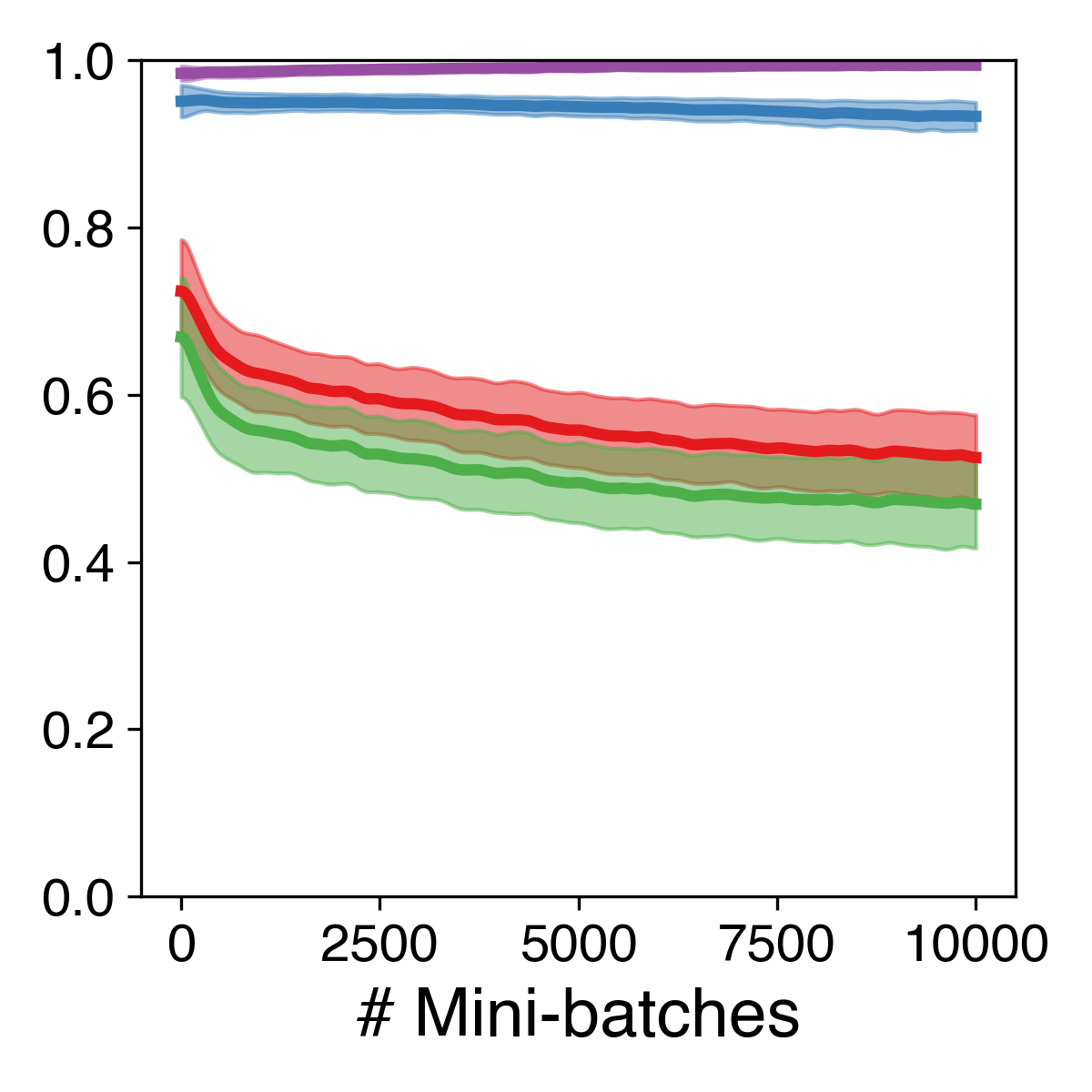}\hfill}
\subfloat[]{
\includegraphics[width=.33\textwidth]{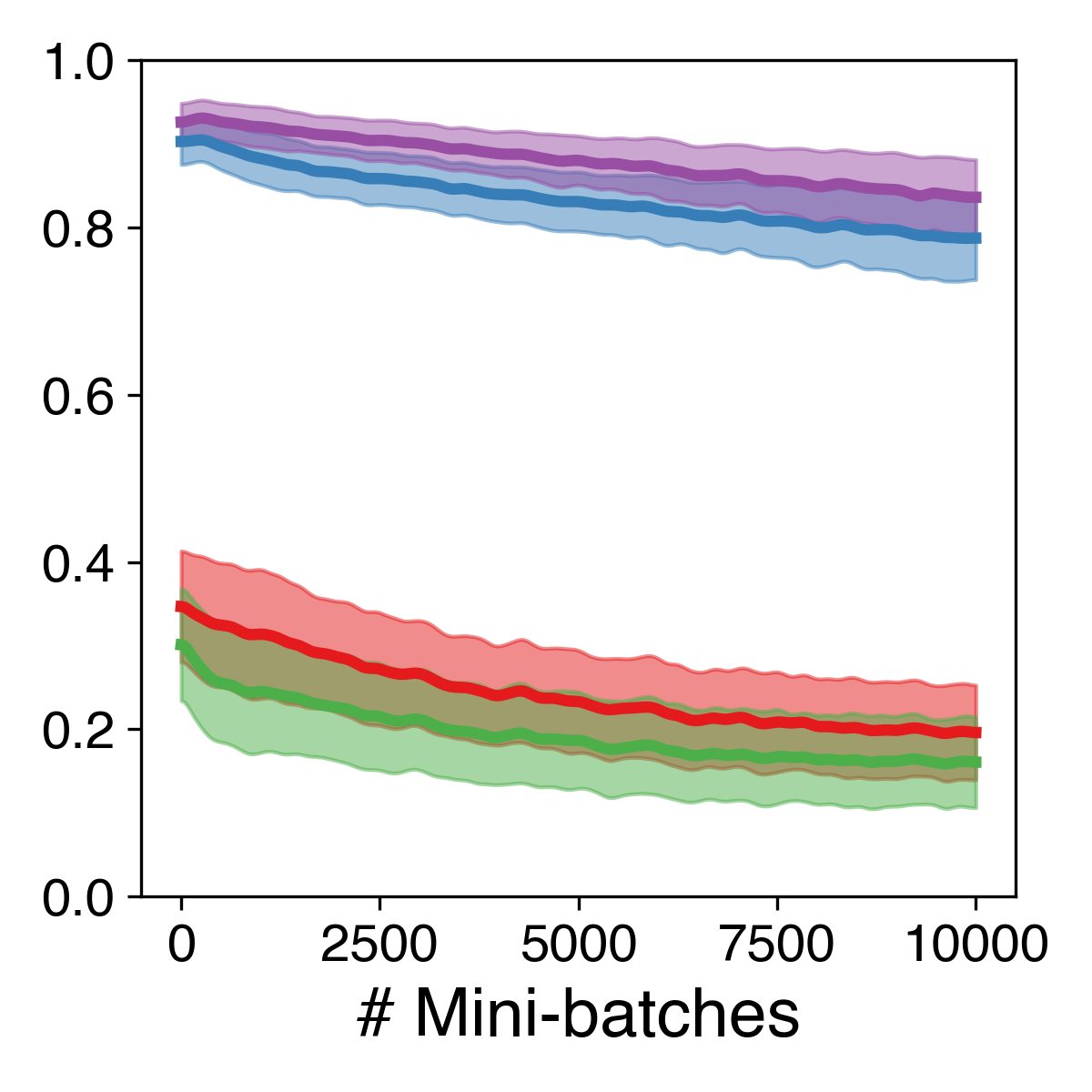}\hfill}
\caption{Mean layer-wise gradient cosine similarity across four seeds throughout offline training on SHD. Figures (a), (b), and (c) are the output layer, second hidden layer, and first hidden layer of 512 layer width, respectively.}
\label{fig:lcos_per}
\end{figure}

\begin{figure}[!htp]

\centering
\subfloat[]{
\includegraphics[width=.34\textwidth]{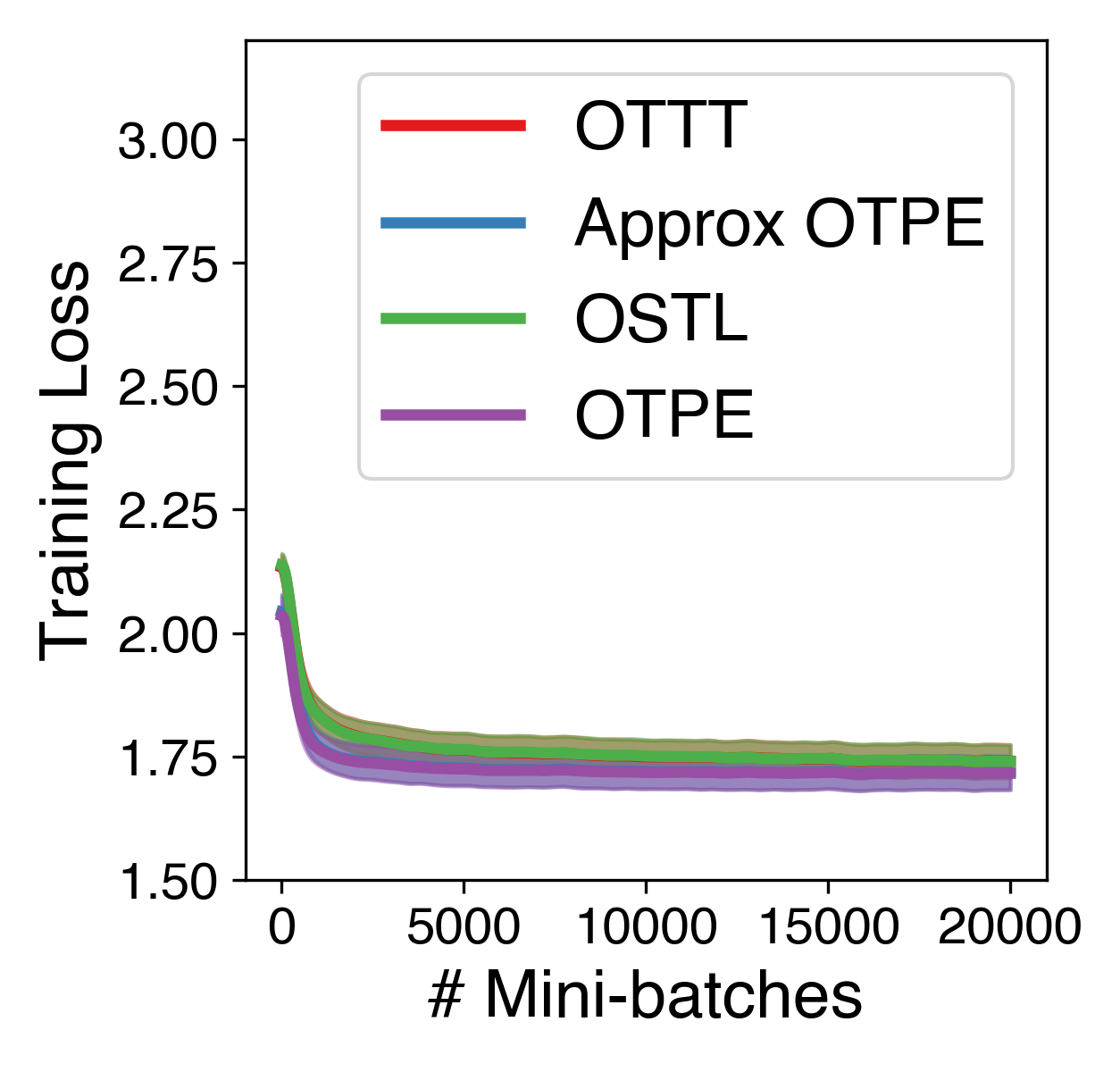}\hfill}
\subfloat[]{
\includegraphics[width=.33\textwidth]{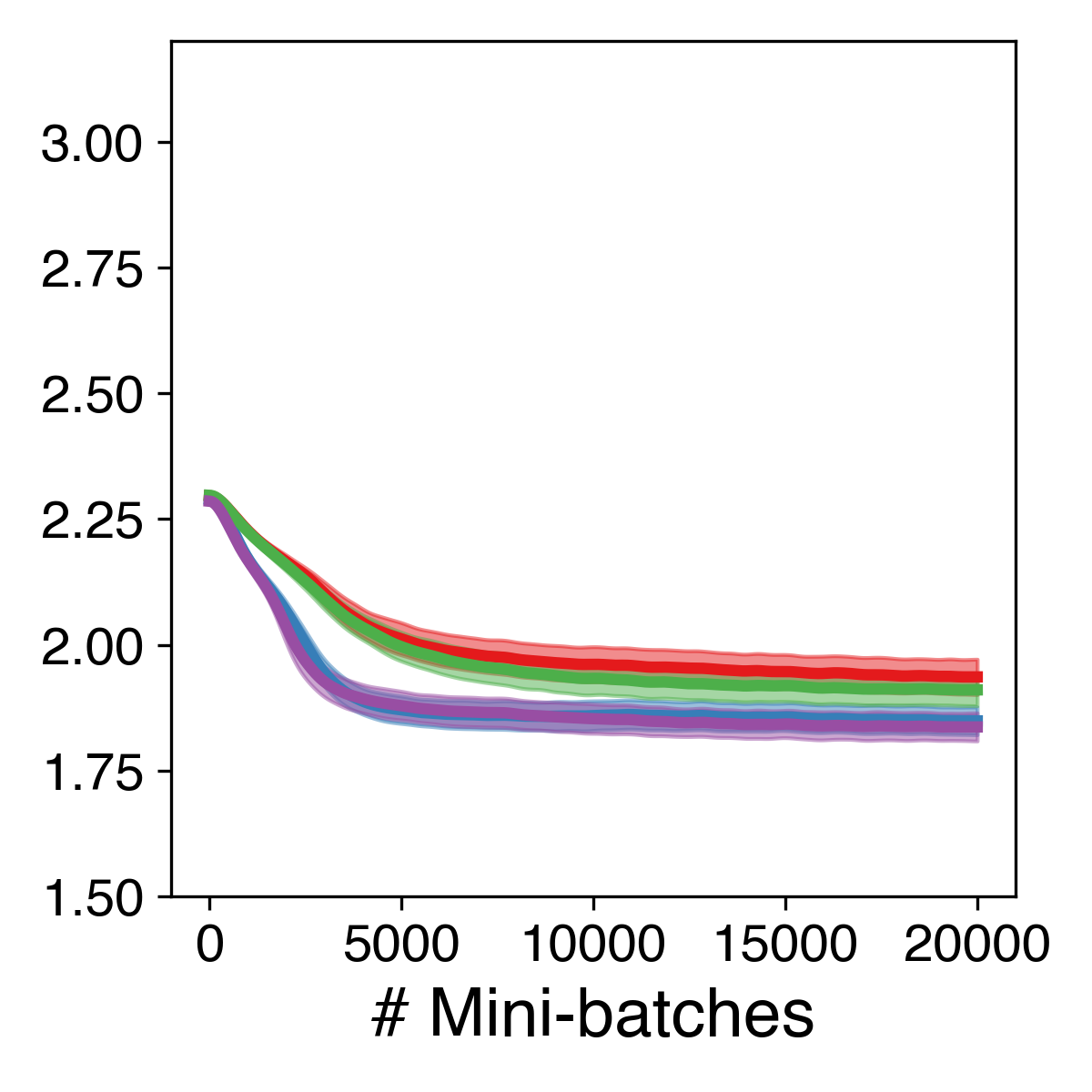}\hfill}
\subfloat[]{
\includegraphics[width=.33\textwidth]{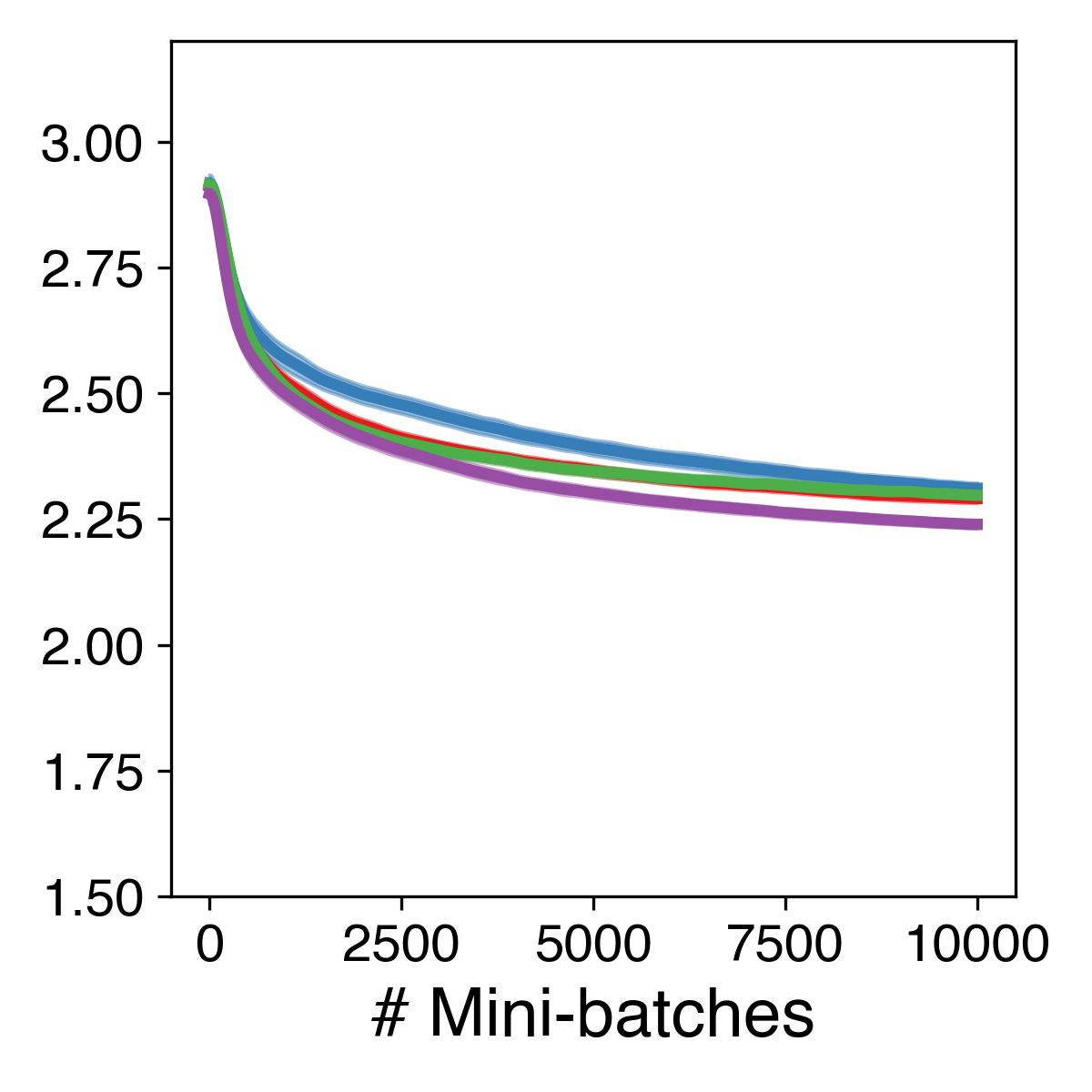}}

\caption{Mean training loss across four seeds throughout offline training. Figures (a), (b), and (c) are R-Randman, T-Randman, and SHD, respectively. Shaded regions are one standard deviation.}
\label{fig:online_loss}

\end{figure}

\begin{figure}[htp]

\centering
\subfloat[]{
\includegraphics[width=.5\textwidth]{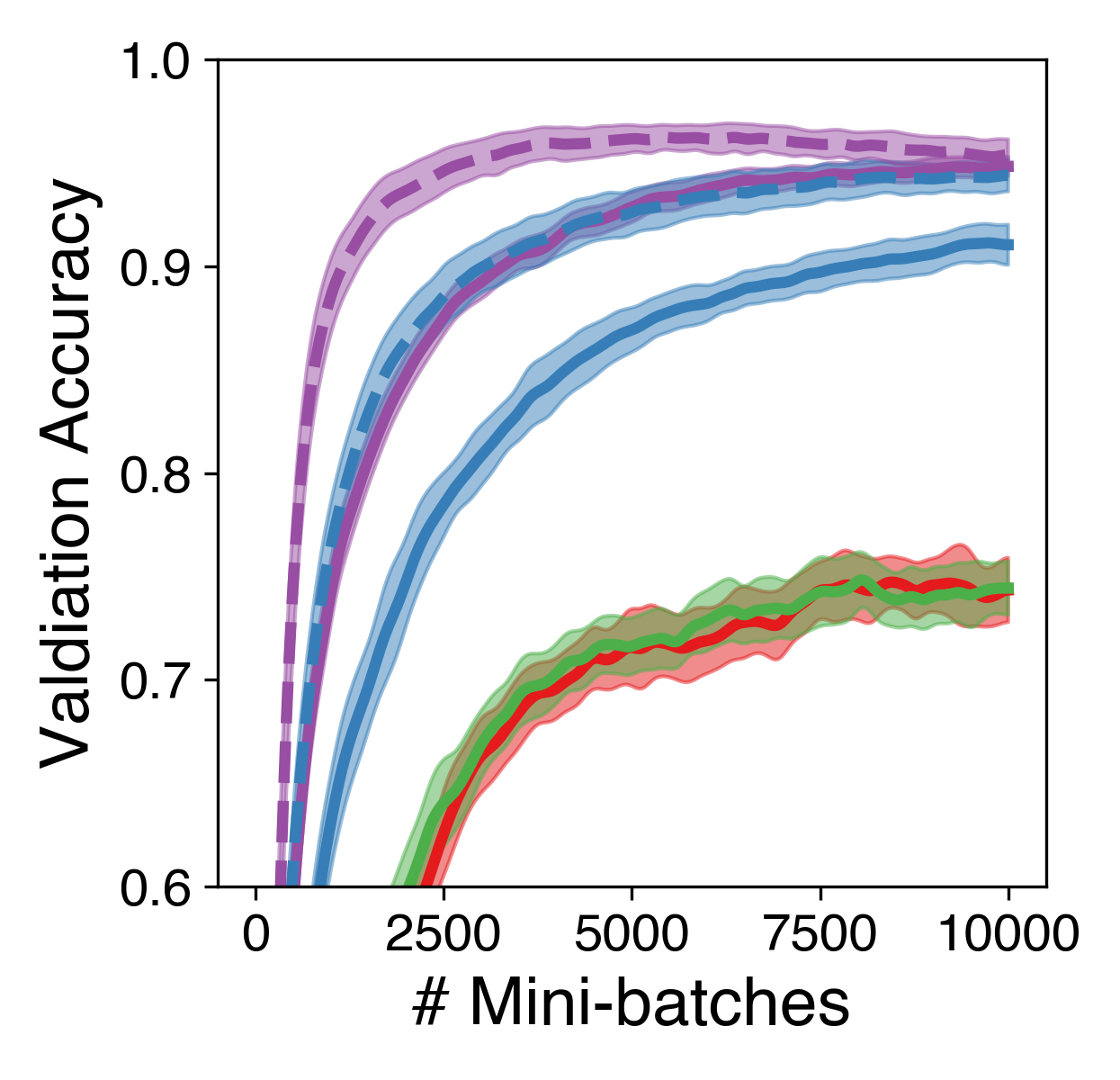}\hfill}
\subfloat[]{
\includegraphics[width=.5\textwidth]{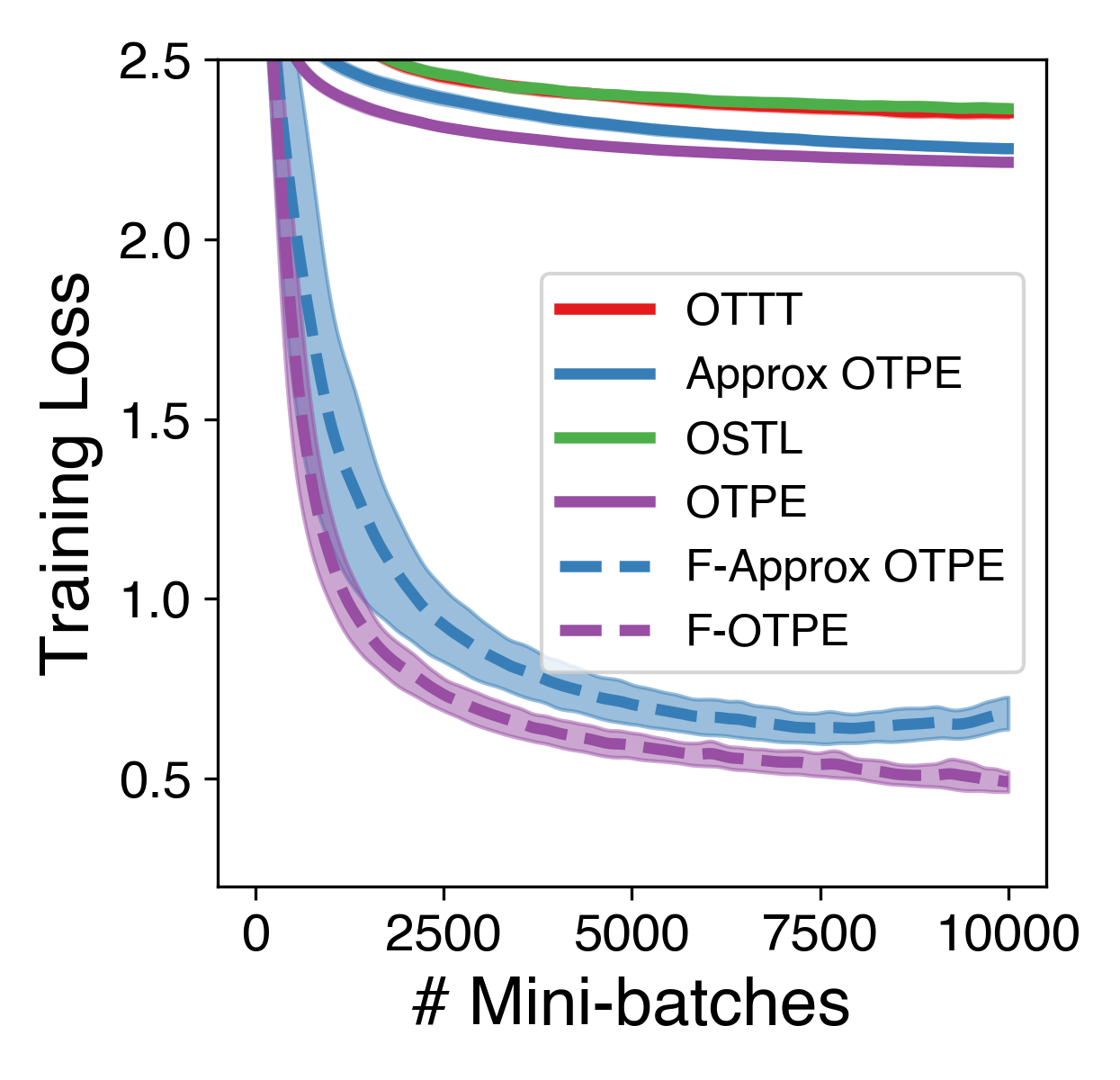}}
\caption{Mean training loss across four seeds throughout offline training. Figures (a), (b), and (c) are R-Randman, T-Randman, and SHD, respectively. Shaded regions are one standard deviation.}
\label{fig:overfit}

\end{figure}

\begin{figure}[H]%[htp]

\centering
\includegraphics[width=.5\textwidth]{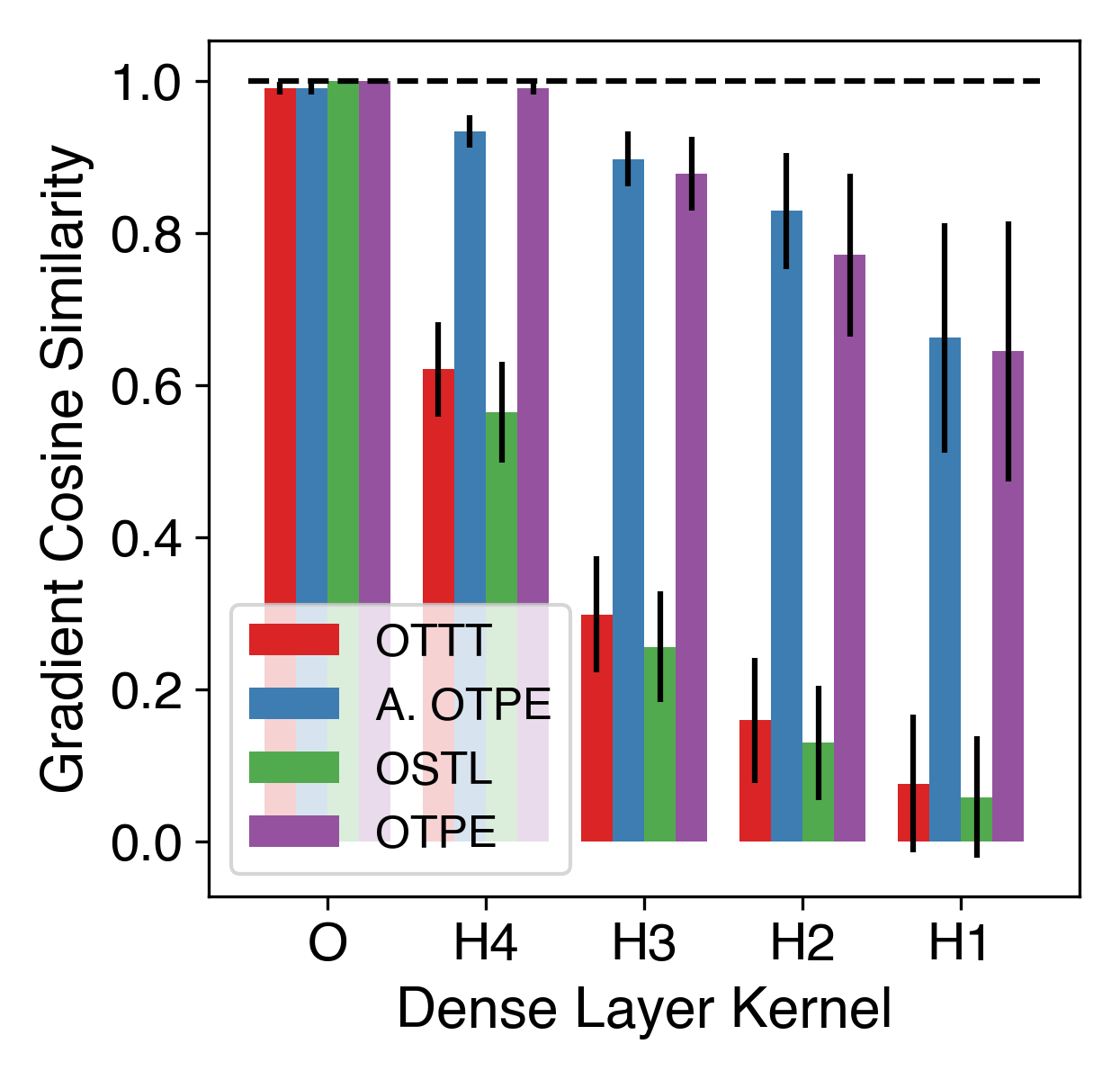}\caption{Gradient cosine similarity for a 5-layer model evaluated on SHD. The error bars are one standard deviation across four seeds and all 10,000 measurements throughout training. In the first hidden layer, gradient cosine similarity is sometimes negative for OSTL and OTTT, which is reflected in the error bars spanning below zero.}\label{fig:5_lcos}
\end{figure}

We additionally include training curves evaluated over different model configurations and hyperparameters in Figs.\ref{fig:lr_search} -- ~\ref{fig:5_lcos}. While the mean gradient cosine similarity for each layer shows a decrease in cosine similarity in earlier layers, Fig. \ref{fig:lcos_per} shows the gradient cosine similarity changing throughout training. Notably, OTTT's cosine similarity increases throughout training in the output layer while both OSTL and OTTT decrease throughout training in the hidden layers. OTPE, on the other hand, increases in the last hidden layer but decreases in earlier hidden layers. In Fig. \ref{fig:overfit}, OTPE decreases in validation accuracy after training on around 5,000 mini-batches. Meanwhile, the loss continues to decrease, which indicates overfitting.

\end{document}